%% file: main.tex
\documentclass{egpubl}
\usepackage{eg2025}
\STAR                   %

\CGFStandardLicense

\usepackage[T1]{fontenc}
\usepackage{dfadobe}  

\biberVersion
\BibtexOrBiblatex
\usepackage[backend=biber,bibstyle=EG,citestyle=alphabetic,backref=true,mincitenames=1,maxcitenames=2,natbib=true]{biblatex} 
\renewcommand*{\mkbibnamefamily}[1]{#1}
\renewcommand*{\mkbibnamegiven}[1]{#1}
\electronicVersion
\PrintedOrElectronic
\ifpdf \usepackage[pdftex]{graphicx} \pdfcompresslevel=9
\else \usepackage[dvips]{graphicx} \fi

\usepackage{egweblnk}

\title[Survey on Modeling of Human-made Articulated Objects]{Survey on Modeling of Human-made Articulated Objects}

\author[J. Liu, M. Savva, A. Mahdavi-Amiri]{
Jiayi Liu \quad Manolis Savva \quad Ali Mahdavi-Amiri\vspace{-5pt}\\
Simon Fraser University
}

\input{macro}
\begin{document}

\teaser{
 \includegraphics[width=\linewidth]{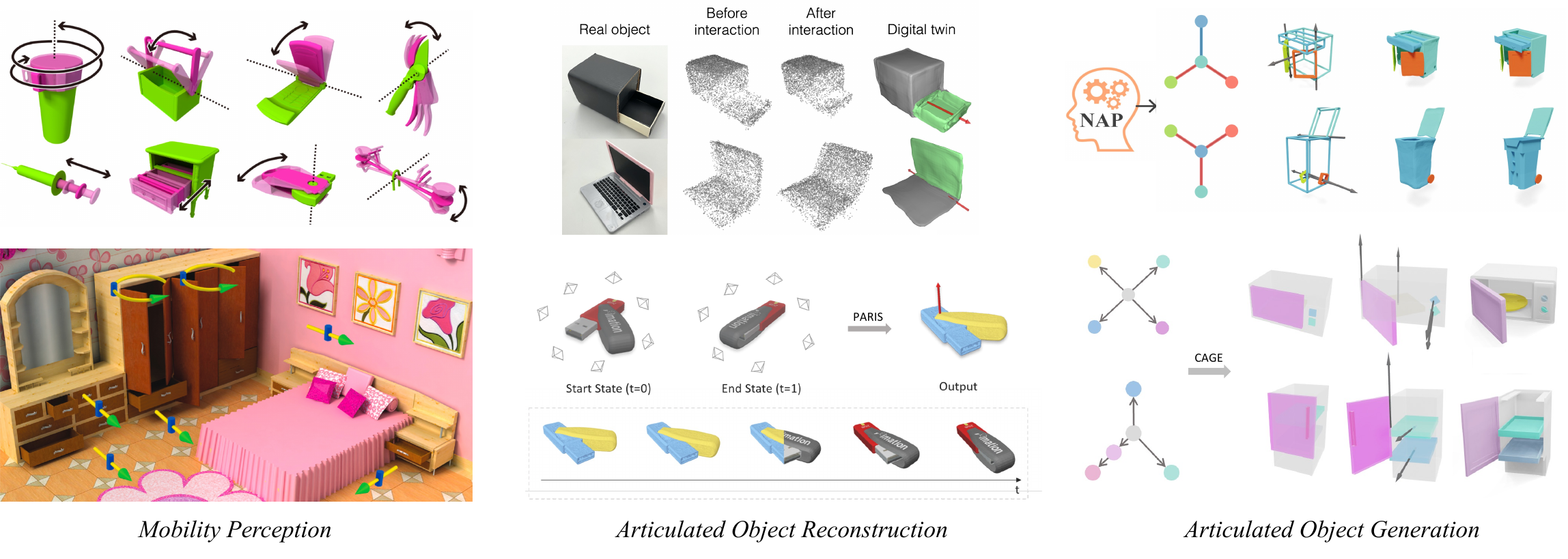}
 \centering
  \caption{
  In 3D modeling of human-made articulated objects, mobility perception, articulated object reconstruction and articulated object generation are common problems that can be solved using a range of approaches and under different settings of input, output, representation, etc.
  From left to right: 
  1) mobility perception from a single snapshot of an object~\cite{hu2017learning} or a scene~\cite{sharf2014mobility}; 
  2) articulated object reconstruction from point clouds~\cite{jiang2022ditto} and multi-view images~\cite{liu2023paris}; 
  3) articulated object generation unconditionally~\cite{lei2023nap} and constrained by a graph~\cite{liu2024cage}.
  Figures reproduced from original papers~\cite{sharf2014mobility,hu2017learning,jiang2022ditto,liu2023paris,lei2023nap,liu2024cage}.
  }
\label{fig:teaser}
}

\maketitle
\begin{abstract}
   3D modeling of articulated objects is a research problem within computer vision, graphics, and robotics. 
   Its objective is to understand the shape and motion of the articulated components, represent the geometry and mobility of object parts, and create realistic models that reflect articulated objects in the real world. 
   This survey provides a comprehensive overview of the current state-of-the-art in 3D modeling of articulated objects, with a specific focus on the task of articulated part perception and articulated object creation (reconstruction and generation). 
   We systematically review and discuss the relevant literature from two perspectives: geometry modeling (i.e., structure and shape of articulated parts) and articulation modeling (i.e., dynamics and motion of parts). 
   Through this survey, we highlight the substantial progress made in these areas, outline the ongoing challenges, and identify gaps for future research. 
   Our survey aims to serve as a foundational reference for researchers and practitioners in computer vision and graphics, offering insights into the complexities of articulated object modeling.
\end{abstract}

\input{secs/1_intro}
\input{secs/2_bg}

\input{secs/3_dataset}

\input{secs/4_geometric}

\input{secs/5_articulation}
\input{secs/6_conclusion}

\vspace{20pt}
\mypara{Acknowledgments.}
This work was funded in part by NSERC Discovery grants (RGPIN-06489-2019, RGPIN-2022-03111), a Canada Research Chair grant (CRC-2019-00298), and a startup grant from Simon Fraser University (N000449).
We thank Angel X. Chang, Richard Zhang, and Han-hung Lee for their helpful comments and discussions. 

\renewcommand*{\mkbibnamefamily}[1]{\textsc{#1}}
\renewcommand*{\mkbibnamegiven}[1]{\textsc{#1}}
\printbibliography

\end{document}

%% file: macro.tex
\usepackage{times}
\usepackage{epsfig}
\usepackage{graphicx}
\usepackage{amsmath}
\usepackage{amssymb}
\usepackage{booktabs}
\usepackage{multirow}
\usepackage{comment}
\usepackage{xcolor}
\usepackage{comment}
\usepackage{cleveref}
\usepackage{graphicx}
\definecolor{mygreen}{RGB}{0, 150, 0}

\newif\ifshowcomments
\showcommentsfalse %

\ifshowcomments
    
    \newcommand{\jy}[1]{\textcolor{purple}{#1}}

\else
    
    \newcommand{\jy}[1]{#1}

\fi

\newcommand{\mypara}[1]{\noindent\textbf{#1}}

\usepackage{pifont}%
\newcommand{\cmark}{\ding{51}}%
\newcommand{\xmark}{\ding{55}}%

%% file: secs/1_intro.tex
\section{Introduction}
Articulated objects, composed of multiple rigid parts connected by joints, are ubiquitous in our daily life, encompassing a wide range of objects, such as mechanical assemblies, furniture pieces, even human bodies and other animals.
Modeling these objects is a research field intersecting computer vision, graphics, and robotics.
It aims to comprehend and represent the shape and mobility of articulated components. 
This endeavor extends to creating 3D models that realistically reflect real-world articulated objects.
Contributing to the dynamics and functionality in our physical world, articulated object modeling is essential for a wide range of applications, such as animation~\cite{yang2022banmo,chen2023fast,qian20233dgs,li2023animatable} and simulation~\cite{weng2019photo,yang2022object,xiang2020sapien}, robotic interaction and manipulation~\cite{hausman2015active,gadre2021act,mo2021where2act,qian2023understanding}, embodied AI~\cite{kolve2017ai2,puig2018virtualhome,gao2019vrkitchen,savva2019habitat,puig2023habitat}, etc.

The complexity of articulated object modeling stems from the fact that articulated objects are not just represented by the geometry of their parts but also by their kinematic structure. 
Over the past decade, much research has focused on addressing these complexities in two main directions: \textit{articulated part perception}, and \textit{articulated object creation}.
 The primary goal of articulated part perception is to analyze the mobility of parts in an object, which is useful in two application areas: 1) object manipulation in robotics; and 2) interaction and animation in simulations.
The vision and robotics communities have been actively working on the first application area~\cite{gadre2021act,mo2021where2act,eisner2022flowbot3d,xu2022universal,wang2022adaafford,mittal2022articulated,geng2023partmanip,bao2023dexart,hsu2023ditto,ning2024where2explore,ling2024articulated,yu2024gamma,wang2024rpmart}, where the task is usually decomposed to affordance prediction, articulation estimation, and action planning.
We leave this line of work out of the scope of this survey, as it is focused on manipulation of articulated objects with robots in the real world.
Another line of work from vision and graphics community focuses more on the geometric and kinematic modeling of the articulated objects for the second application area, where the task is usually decomposed to articulated part segmentation and kinematic structure understanding.
For articulated object creation, the main goal is to reconstruct or generate articulated parts into a compositional and hierarchical 3D model that represents a real-world object that can be interacted with in a virtual environment.

This survey provides a comprehensive overview of the current state-of-the-art in 3D modeling of human-made articulated objects.
Through this survey, we highlight the substantial progress made in these areas, outline the ongoing challenges, and identify the gaps for future research. 
Our survey aims to serve as a foundational reference for researchers and practitioners in computer vision and graphics, offering insights into the complexities of articulated object modeling and inspiring new research in this area.

We begin by laying the groundwork with an overview of the field, defining the scope and focus of our discussion in~\Cref{sec:bg}.
Next, in~\Cref{sec:dataset}, we compile the key datasets for articulated objects that have been collected and utilized within the research community, highlighting their critical role in advancing the 3D modeling of articulated objects.
Our survey then delves into an in-depth analysis of the techniques and methodologies developed for 3D modeling of articulated objects.
This analysis is organized around two pivotal axes: geometry  modeling and articulation modeling, which are the foundational components to tackle the complexities in this field.
Geometry modeling covers the representation and approaches employed to understand the shape of the articulated parts, whereas articulation modeling involves the articulation model and the methodologies used to estimate the part mobility and kinematic structure of the object.
In the end, we conclude our survey by identifying the key challenges and potential research directions that can drive future progress in the field.

\mypara{Related surveys.}
The most related survey to ours is the recent survey by \citet{pejic2022articulated}, which focuses on the manipulation of articulated objects in the realm of robotics and computer vision.
In contrast, our focus is on the geometry and articulation modeling of articulated objects within the field of computer vision and graphics.
This aspect represents a distinct yet critical dimension of articulated object research, orthogonal to the topic of manipulation.
Additionally, another related survey by \citet{hu2018functionality} explores the functionality of general objects within the scope of computer graphics literature. 
The functionality of an object is influenced not just by its mobility but also by its shape, material, and various other attributes that collectively determine its capability to perform a task.
Our survey takes a more specialized focus on the articulated object and its mobility particularly, compared to the work by \citet{hu2018functionality} which concentrates on functionality for general objects.

%% file: secs/2_bg.tex
\input{figures/2_joint_types.tex}

\section{Background \& Scope}
\label{sec:bg}

\subsection{Definition of Articulated Objects}
\label{subsec:bg_def}
An articulated object is composed of multiple rigid parts interconnected by joints.
These joints serve as the pivotal points that allow for relative motion between the connected parts.
The range and nature of the articulation of each part are constrained and defined by the type of joint it possesses. 
Among human-made objects, the most frequently encountered joint types facilitate linear motion, such as revolute, prismatic, and helical joints.
For example, a laptop is an articulated object with a revolute joint connecting the screen to the keyboard.
This joint allows the screen to rotate around a single axis relative to the keyboard, with the range of motion being restricted to a specific limit.
In contrast, the human body is a more complex example of an articulated object, with a wider variety of joint types, such as ball-and-socket or spherical joints.
These joints allow for motion beyond linear translation or rotation, enabling a more diverse range of movements.
The illustration of several common joint types is shown in \Cref{fig:joint_types}.

\mypara{Types of articulated objects.} 
Articulated objects, with their distinctive characteristics and varied types, play a significant role in the dynamics of our living environments.
In general, articulated objects can be classified into two types: \emph{organic objects} and \emph{human-made objects}.

\begin{itemize}
    \item \textbf{Organic objects} refer to naturally occurring systems that are capable of initiating movement through articulated structures. 
    The bodies of humans and other animals are common examples of organic articulated objects. 
    They are crucial for various functions such as locomotion, manipulation, and interaction with the environment.
    A skeleton with bones and joints is usually used to represent the kinematic structure of these objects.
    The motion allowed for these objects is usually structurally complex with a relatively large number of joints varied in the topology of connections and degrees of freedom.
    However, the kinematic structure is relatively fixed and can be predefined for a specific species.
    How to extract effective skeletons, reconstruct 3D models, and fit dynamic motions for animals across various species is an active research topic~\cite{kulkarni2020articulation,yang2021lasr,wu2022casa,yao2022lassie,wu2023magicpony,aygun2023saor,lei2023gart,yao2023hi,li2024learning,yao2024artic3d,liu2024lepard,jakab2024farm3d,maheshwari2023transfer4d,yang2024omnimotiongpt}.
    Leveraging the skeleton template of the human body or hands and coordination between the joints to achieve the desired animation~\cite{aberman2020skeleton,su2021nerf,bergman2022generative,su2022danbo,wang2022arah,kuai2023camm,lei2023gart,zhang2023generating,li2023animatable,qian20233dgs,tan2023distilling} or human-object interaction~\cite{taheri2020grab,zhang2020perceiving,xu2021d3d,cao2021reconstructing,zhang2022couch,fan2022articulated,bhatnagar2022behave,qian2022understanding,ghosh2023imos,jiang2023full,huang2023one,lee2023locomotion,xie2024rhobin} is another line of research. 
    We direct readers to a recent survey on human body modeling~\cite{berretti2018representation,chen2021towards,wang2021deep,mourot2022survey,akber2023deep,zheng2023deep,yang2024digital} and 3D hand modeling~\cite{cheng2015survey,wheatland2015state,barsoum2016articulated,li2019survey,ahmad2019hand} for more details.
    In this survey, we leave the discussion of organic articulated objects out of the scope and only focus on human-made articulated objects.
    \item \textbf{Human-made objects} are engineered and designed to mimic the natural articulation of organic objects or to serve specific functionalities that require movement.
    These objects are complex assemblies of rigid parts connected by joints, and their motion is often a direct result of interaction with the environment or driven by an active system.
    Common examples of human-made articulated objects include gadgets, furniture, vehicles, and other mechanical systems.
    These objects are ubiquitous in our daily lives and play an important role in our interaction with the physical world.
    Modeling human-made articulated objects presents several unique challenges in terms of physical and geometric modeling, articulation modeling, and environmental interaction. 
    Modeling these objects is the focus of this survey and will be discussed in detail in the following sections.
    We will only mention articulated objects for short in the rest of the survey to refer to human-made articulated objects.
\end{itemize}

\mypara{\jy{Key differences.}} 
\jy{
Modeling organic and human-made articulated objects presents distinct challenges due to differences in geometry, kinematic structure, motion patterns, and interaction with the environment.
Topologically, organic objects share consistent skeletal structures within species, enabling predefined templates (e.g., human body, quadrupeds), whereas human-made objects exhibit greater variability.
Some categories follow standard designs (e.g., eyeglasses, scissors), while others (e.g., storage furniture) vary significantly across instances.
In terms of motion, organic objects have biomechanical joints driven by muscle forces, leading to soft-tissue deformations, whereas human-made objects typically follow rigid-body mechanics with constrained degrees of freedom.
While this generally simplifies articulation modeling, challenges arise in accurately estimating joint parameters in relation to object geometry, handling nested joints with complex kinematic dependencies (e.g., bi-fold doors, folding umbrellas), and accounting for real-world imperfections such as frictional resistance and mechanical backlash.
Additionally, organic objects evolve for locomotion and behaviors, while human-made objects are designed for specific functionalities, influencing how they articulate and are manipulated.
Consequently, research on organic objects emphasizes geometric reconstruction and animation, whereas human-made objects are studied in the context of 3D modeling for robotic manipulation and functional interaction.
}

\jy{
These fundamental differences make it difficult for methods developed for one type to directly generalize to the other.
Structure-wise, the parametric models and statistical priors used for organic objects rely on consistent skeletal structures, which are not scalable to accommodate the diverse topologies of human-made objects.
Geometrically, organic objects are typically represented as a single continuous surface, i.e., skin enclosing a skeleton.
Reconstruction methods optimized for organic shapes rely on this assumption and are not well-suited for human-made objects, which often consist of multiple discrete, interlocking parts with internal structures, requiring different modeling approaches.
Articulation-wise, organic objects exhibit complex, non-linear motion patterns driven by muscle forces, necessitating biomechanical models and motion capture data for accurate representation.
On the other hand, the rigid-body mechanics of human-made objects are more amenable to analytical kinematic models, simplifying articulation modeling but requiring precise joint parameter estimation to enable plausible interaction in physical simulations.
}

\subsection{Representation of Articulated Objects}
Articulated object modeling is a multifaceted domain within computer vision, graphics, and robotics.
It aims to understand the shape and interactability of the articulated components, represent the geometry and mobility of these objects, and create realistic models that reflect articulated objects in the real world.
The inherent complexity of this task arises from the fact that the articulation of the object is not only determined by the geometry of the parts but also by the kinematic structure of the object.
As a result, effective modeling of articulated objects requires capturing a dual representation of the geometry and articulation of the object.
These two facets are deeply interconnected, each influencing and informing the other. 
The geometric decomposition of the parts lays the foundational groundwork for analyzing part mobility, while the process of articulation modeling also guides the shape understanding of the mobility parts.
This symbiotic relationship between geometric and articulation modeling underscores the importance of a cohesive approach in articulated object modeling, where understanding and representing both geometry and articulation are essential for capturing the full essence and functionality of complex entities.
Over the last decade, there has been a notable surge in research within this field, addressing the inherent challenges predominantly from two perspectives: \textit{geometry modeling} and \textit{articulation modeling}.
We will further discuss these two axes in the following sections.
\input{figures/2_geo_rep.tex}

\mypara{Geometric representation.}
Unlike the geometric modeling of static objects, which typically concentrates solely on the object's outer surface, the geometric representation of articulated objects demands a more nuanced approach. 
This involves capturing the spatial arrangement of the parts, modeling the part-level surfaces, and constructing the interior structure of the object.
It is essential to accurately describe the shape, size, relative position, and orientation of each individual part that is structurally organized within the object. 
This comprehensive description is crucial, as it provides the foundational details necessary to represent a complete object that can be interactively manipulated.
The data format used to represent object geometry varies based on the data capture methods and the intended application.
The most common access to the part geometry is through the 3D point cloud, which is a set of points sampled from the object surface that can be obtained by 3D scanning or depth projection.
While point clouds provide a raw, unstructured representation of the object's geometry, they often undergo further processing into more organized formats for detailed analysis.
Another common representation is 3D mesh, which approximates the surface of each part  with primitives such as polygons.
 The mesh format is widely used in applications such as simulation, rendering, and animation due to its ability to effectively model surface details and because it can flexibly represent object structure.
In \Cref{sec:geometric}, we will discuss different choices of geometric representation in various related tasks in detail.
Here we specify the geometric representations that are considered in the literature and visually illustrated in \Cref{fig:geo_rep}.
\begin{itemize}
    \item \textbf{Mesh}: is a collection of vertices, edges, and polygons usually with a structured connectivity (e.g., 2-manifold) that approximate the surface.
    \item \textbf{Point cloud (PC)}: a set of 3D points sampled on object surface that can be obtained from 3D scanning or depth projection.
    \item \textbf{Implicit field}: is an implicit function that assigns a value to each point in the space to construct a field describing the object. 
    Common forms of implicit fields include the signed distance field (SDF), occupancy field, density field, etc.
    The function parameterized by a neural network, known as a neural field, is a popular intermediate representation as it is inherently differentiable and easy to integrate with learning-based methods.
    \item \textbf{ Images-based}: some work approaches the 3D modeling  and perception problem from the image-based domain, such as depth images, RGB-D images and posed RGB images captured from multi-views or stereo cameras.
\end{itemize}

\input{figures/2_arti_rep.tex}

\mypara{Articulation representation.}
The representation for articulation includes the description of the mobility of each part and the kinematic structure of the object.
The kinematic structure describes how the object's parts are connected and how they can move relative to each other. 
Typically, this kinematic structure is represented by a tree or graph, which is an effective way to model hierarchical relationships in articulated systems.
The part mobility can be described by the parameters of each connected joint or the deformation field of the object.
In \Cref{sec:articulation}, we will discuss ways of articulation modeling with different assumptions and representation in various related tasks in detail.
Here we specify the articulated motion representation as follows and visually illustrated in \Cref{fig:arti_rep}.
\begin{itemize}
    \item \textbf{Deformation field}: the deformation field is a general form of motion representation that describes the displacement of each point as a 3D vector. It can also be used to represent beyond the rigid motion of the shape in a continuous space, such as free-form deformation.
    \item \textbf{Joint parameters}: the parameters of each joint, including the \emph{joint type}, the \emph{joint axis}, the \emph{joint state}  such as the rotational angle or translational distance, and the \emph{motion range} or \emph{joint limit}. 
    \item \textbf{Kinematic tree}: the kinematic structure of the object represented by a tree or graph, where the nodes represent the articulated parts and the edges represent the joint connections.
\end{itemize}

\input{figures/3_datasets.tex}

\input{tabs/dataset}

%% file: figures/2_joint_types.tex
\begin{figure}[t]
    \begin{center}
    \includegraphics[width=\linewidth]{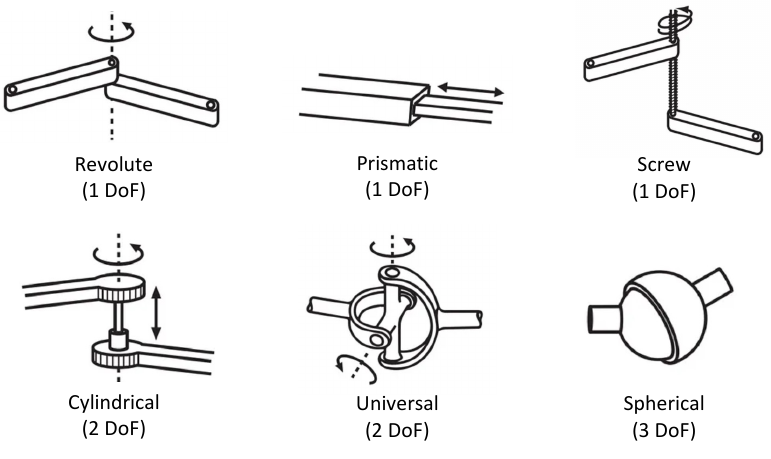}
    \end{center}
    \caption{Common joint types. 
    The first row shows the joints with linear motion which is the most common assumption considered in the literature.
    The second row shows more complex joints with more than one DoF.
    Figure reproduced from original paper~\cite{lynchpark2017modern}.}
    \label{fig:joint_types}
\end{figure}

%% file: figures/2_geo_rep.tex
\begin{figure}[t]
    \begin{center}
    \includegraphics[width=\linewidth]{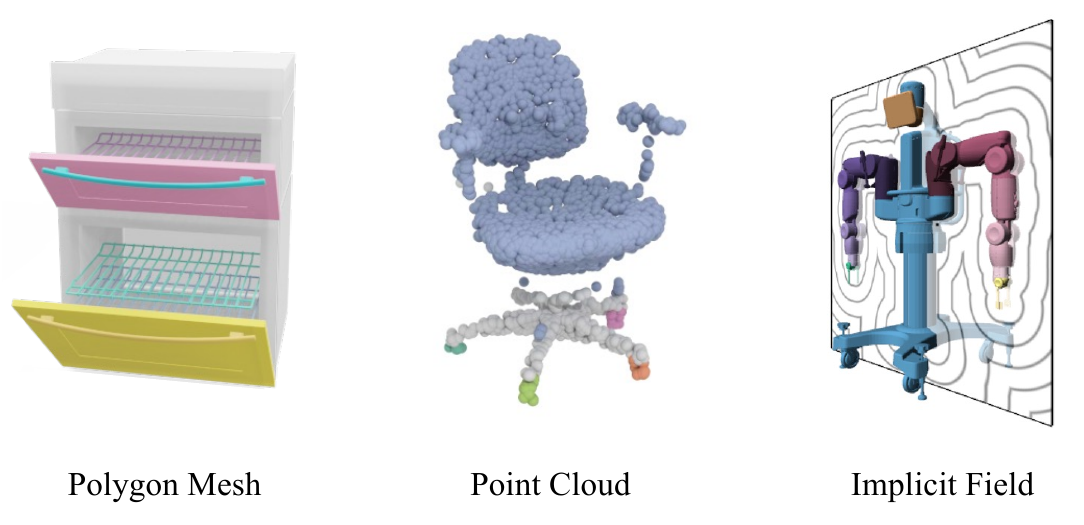}
    \end{center}
    \caption{Common geometric representations for articulated object modeling.
    Figures reproduced from original papers~\cite{liu2024cage,yan2019rpm,schmidt2014dart}.
    }
    \label{fig:geo_rep}
\end{figure}

%% file: figures/2_arti_rep.tex
\begin{figure}[t]
    \begin{center}
    \includegraphics[width=\linewidth]{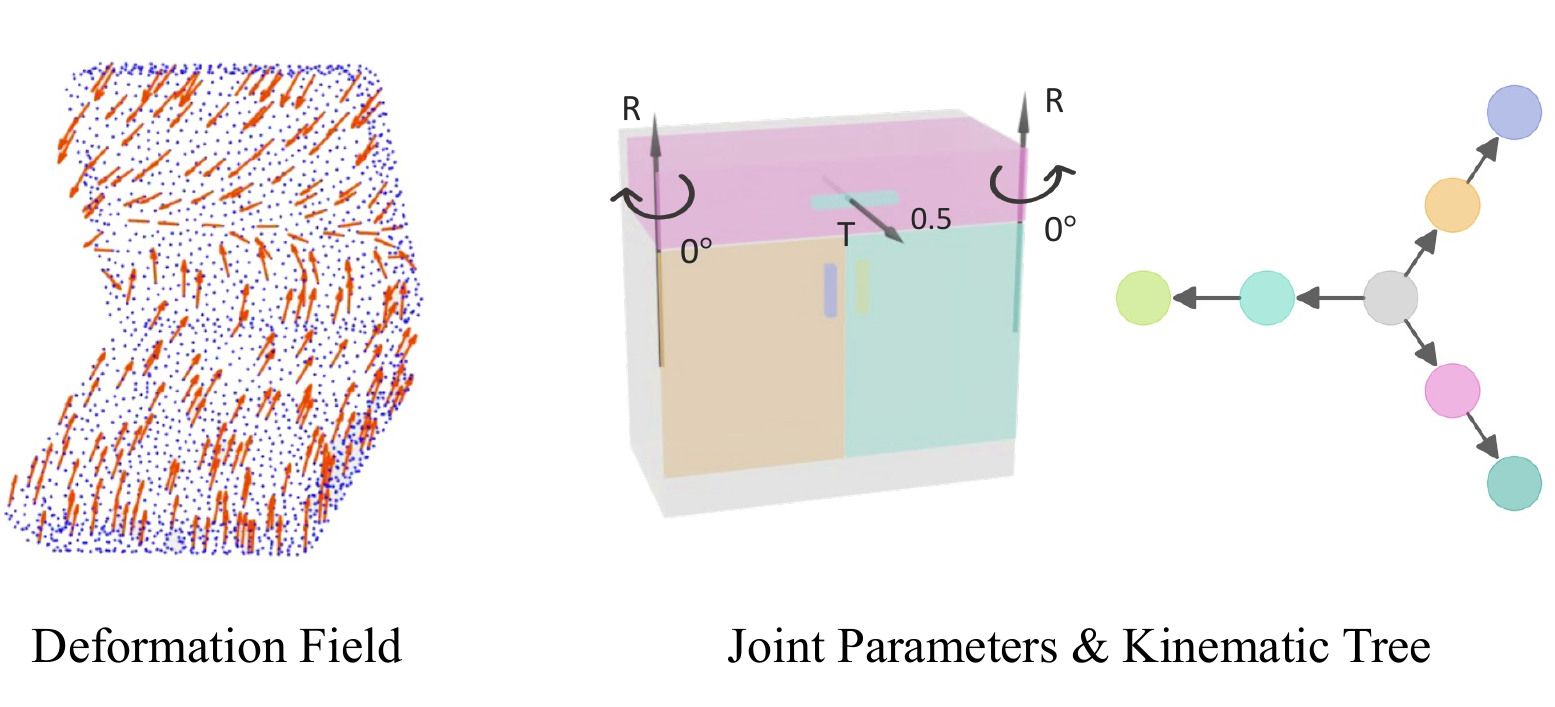}
    \end{center}
    \caption{Common articulated motion representations. 
    Figures reproduced from original papers~\cite{yi2018deep,liu2024singapo}.}
    \label{fig:arti_rep}
\end{figure}

%% file: figures/3_datasets.tex
\begin{figure*}[t]
    \begin{center}
    \includegraphics[width=\linewidth]{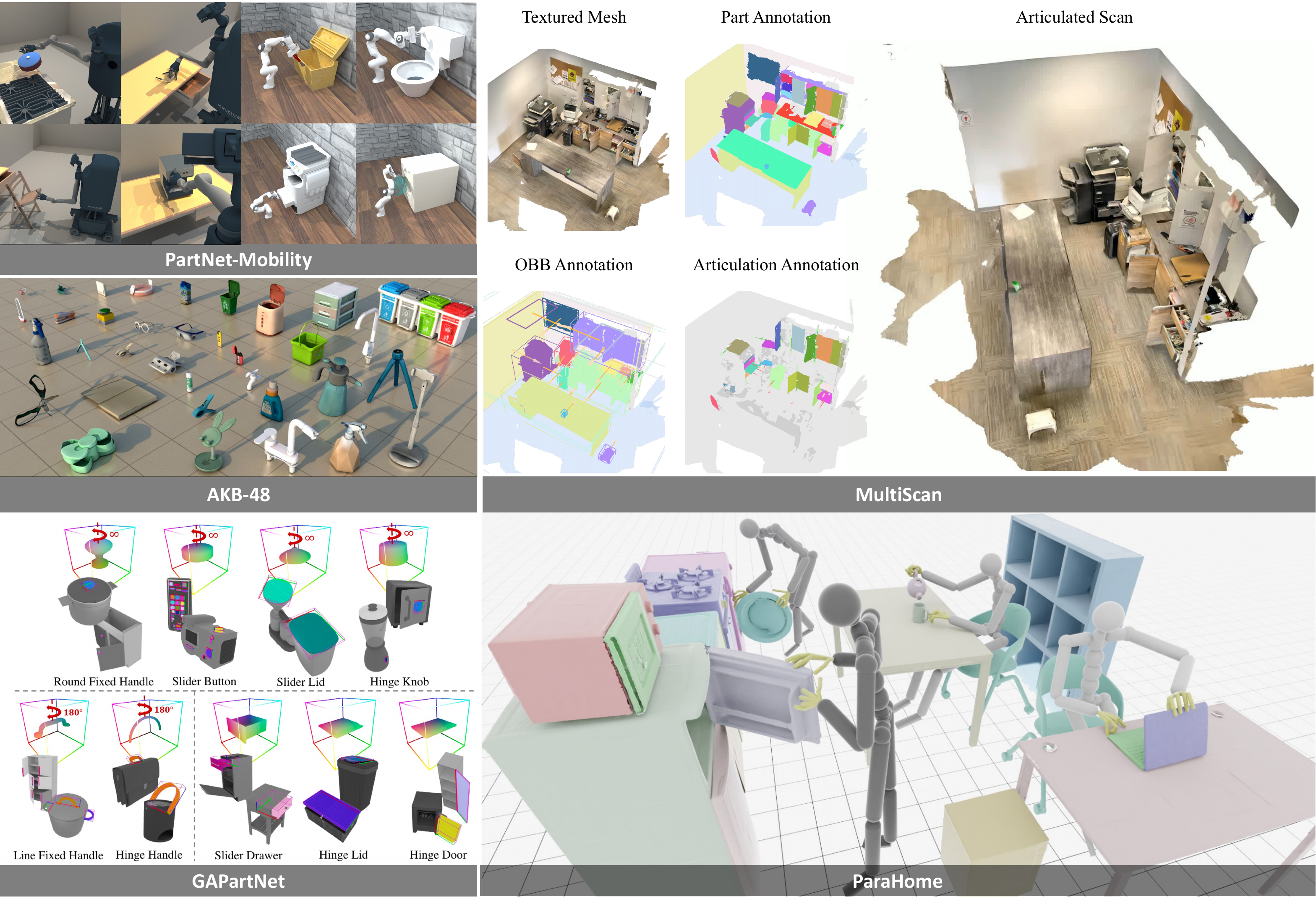}
    \end{center}
    \caption{Representative datasets for 3D articulated objects: PartNet-Mobility~\cite{xiang2020sapien} for synthetic data, AKB-48~\cite{liu2022akb} for real scans of objects, MultiScan~\cite{mao2022multiscan} for real indoor scenes with articulated objects, and GAPartNet~\cite{geng2023gapartnet} for synthetic and real objects with fine-grained annotations for the generalizable actionable parts. ParaHome~\cite{kim2024parahome} is a recent dataset providing human-object interactions with articulated objects.}
    \label{fig:datatsets}
\end{figure*}

%% file: tabs/dataset.tex
\begin{table*}
\centering
\resizebox{\textwidth}{!}{
\begin{tabular}{@{}llllcrrr@{}}
\toprule
                                            & Source            & Level     & Representation    & Textured      & \# objects    & \# movable parts  & \# object categories   \\ \midrule
\citet{hu2017learning}                      & synthetic         & object    & mesh              & \xmark        & 368           & 368               & -  \\
RPM-Net~\cite{yan2019rpm}                   & synthetic         & object    & mesh              & \xmark        & 969           & 1,420             & 43  \\
Shape2Motion~\cite{wang2019shape2motion}    & synthetic         & object    & mesh              & \xmark        & 2,440         & 6,762             & 45 \\
RBO~\cite{martin2019rbo}                    & synthetic, real   & object    & mesh, RGB-D       & \cmark        & 14            & 21                & 14 \\
PartNet-Mobility~\cite{xiang2020sapien}     & synthetic         & object    & mesh              & \cmark        & 2,346         & 14,068            & 46 \\ 
ReArt-48~\cite{liu2022toward}               & real              & object    & mesh              & \cmark        & 48            & -                 & 5  \\ 
AKB-48~\cite{liu2022akb}                    & real              & object    & mesh              & \cmark        & 2,037         & -                 & 48 \\ 
OPDSynth~\cite{jiang2022opd}                & synthetic         & object    & mesh, RGB         & \cmark        & 683           & 1,343             & 11 \\ 
OPDReal~\cite{jiang2022opd}                 & real              & object    & mesh, RGB-D       & \cmark        & 284           & 875               & 8  \\ 
MultiScan~\cite{mao2022multiscan}           & real              & scene     & mesh, RGB-D       & \cmark        & 10,957        & 5,129             & 20 \\ 
OPDMulti~\cite{sun2023opdmulti}             & real              & scene     & RGB-D             & \cmark        & 217           & 688               & 33 \\
GAPartNet~\cite{geng2023gapartnet}          & synthetic, real   & object    & mesh              & \cmark        & 8,489         & 1,166             & 27 \\
ACD~\cite{iliash2024s2o}                    & synthetic         & object    & mesh              & \cmark        & 354           & 1,350             & 21 \\
ParaHome~\cite{kim2024parahome}             & real              & object    & mesh              & \xmark        & 22            & -                 & 8 \\
\bottomrule 
\end{tabular}
}
\caption{
Datasets related to articulated objects with statistics on the source of collection, the level at which these datasets operate, data representation, whether the 3D model is texture, the number of objects, movable parts, and object categories.
The '-' symbol indicates that the information is not available or not reported in the original paper.
}
\label{tab:dataset}
\end{table*}

%% file: secs/3_dataset.tex
\section{Datasets}
\label{sec:dataset}

One of the main contributors to recent advances in deep learning is the availability of large-scale 3D datasets for various computer vision tasks, particularly in shape understanding. 
However, datasets for articulated objects are relatively scarce compared to those for static 3D objects. 
Gathering data for articulated objects is more laborious because it requires detailed modeling of each part's geometry and careful annotation of articulation parameters.
The challenges in creating datasets for articulated objects are multifaceted:
\begin{itemize}
    \item The creation of synthetic datasets involves manual design of the geometry for each component  and annotating the articulation parameters.
    This labor-intensive process restricts the number of categories and instances in these datasets, often leading to a lack of diversity and complexity in the data samples. 
    As a result, models may be less realistic or overly simplistic.
    \item Real-world datasets are typically gathered using sensors like RGB-D cameras, followed by post-processing to reconstruct shapes and annotate part attributes. 
    One of the significant challenges in modeling articulated objects, as opposed to static surface models, is the difficulty in capturing the interior structure of each part from surface data alone. 
     It usually involves multiple scans from different states of the object to capture the complete geometry.
    The data obtained from surface sensors is often incomplete and noisy, as the interior structure frequently is occluded by the outer surface. 
    This limitation affects the geometric quality of the data and further reduces the diversity and complexity of data samples.
\end{itemize}
With the recent surge in interest in articulated objects, there has been a growing effort to create datasets for this domain.
 \Cref{fig:datatsets} shows several representative datasets for 3D articulated objects.
We list all existing datasets for articulated objects in both object-level and scene-level in \Cref{tab:dataset},  where we summarize the key characteristics of each dataset.
 We also provide a brief overview of each dataset in the folllowing subsections.

\subsection{Synthetic Data}
The recent advances in 3D modeling technology have led to a significant increase in the availability of 3D models with part-level structures accessible online. 
This progress has opened up new possibilities for detailed analysis and application in various fields, enhancing the resources for researchers and practitioners working with complex 3D structures.
Articulated objects are one such class of complex 3D structures that have benefited from this progress.
The availability of 3D datasets with part-level structures has enabled the creation of synthetic datasets for articulated objects.

To facilitate the prediction of part mobility via a learning-based approach, \citet{hu2017learning} collect a synthetic dataset for 3D articulated objects by sourcing shapes from ShapeNet~\cite{chang2015shapenet} and SketchUp~\cite{sketchup}.
In this process, they manually segment the shapes into parts and provide detailed annotations for the articulation parameters.
Following this foundational work, RPM-Net~\cite{yan2019rpm} extends the scope of the dataset to include a greater variety of objects incorporating more complex motion, e.g., the opening of the umbrella cover.
For each shape in the dataset, every possible pair of parts is labeled as a \textit{reference part} and a \textit{moving part}. 
Further, each pair is  regarded as a \textit{mobility unit}, associated with annotated motion parameters.
These parameters form a quadruple, including the transformation type (including translation, rotation, or the combination of the two), the position and direction of the transformation axis, and the range of the motion.

Shape2Motion~\cite{wang2019shape2motion} is another synthetic dataset in a larger scale that was introduced concurrently.
The shapes are sourced from ShapeNet~\cite{chang2015shapenet} and 3D Warehouse~\cite{threedwarehouse}.
Each shape includes part segmentations and articulation parameters that are annotated in a similar way as RPM-Net~\cite{yan2019rpm} and \citet{hu2017learning} with the use of a developed annotation tool. 
This tool enhances the efficiency of the annotation process by allowing the annotators to visually verify the correctness by animating the object with the annotated parameters.
This allows for boosting the scale of the dataset with a much larger number of objects and movable parts in a wide range of categories.

The datasets mentioned  above are valuable resources for research on shape analysis and mobility prediction of articulated objects using 3D meshes or point clouds as input.
However, a notable limitation of these datasets is the absence of texture information. 
This restricts their utility in applications such as simulation, digital twins creation, augmented reality, and other modeling tasks that require a more holistic visual representation.
To fill in this gap, the PartNet-Mobility dataset, with a simulation environment SAPIEN ~\cite{xiang2020sapien}, was introduced.
This dataset is a subset and an extension from a part-level 3D shape dataset PartNet~\cite{mo2019partnet}, enriched with articulation annotations organized in URDF files.
The inclusion of diverse appearances and motions in PartNet-Mobility significantly enhances its value, inspiring increasingly more research in the field of articulated object modeling  in connection with visual perception~\cite{jiang2022opd,sun2023opdmulti,wang2025active} and digital twin creation~\cite{jiang2022ditto,liu2023paris,weng2024neural}.
Building upon the PartNet-Mobility dataset, OPDSynth~\cite{jiang2022opd} is introduced for the task of \textit{openable part detection} (OPD) by blending the rendering of the synthetic data with real-world RGB images. 
This dataset enables the mobility part detection from images, aligning more closely with practical real-world applications.

Later, the GAPartNet~\cite{geng2023gapartnet} dataset is introduced to capture finer-level part details that have been previously overlooked in existing datasets.
This dataset is tailored to enhance the generalization of object perception and manipulation tasks across various object categories by focusing on the concept of \textit{generalizable and actionable parts}. 
The underlying premise is that functional parts, such as buttons and handles, are fundamental elements whose identification and extraction can significantly improve generalizability within and across object categories.
They argue that the functional parts such as buttons and handles are more elementary, and the extraction of these parts can improve the generalizability in an intra-category manner.
To achieve a comprehensive and versatile dataset, GAPartNet selectively compiles data from both the PartNet-Mobility~\cite{xiang2020sapien} and AKB-48~\cite{liu2022akb} datasets to encompass both synthetic and real data.
 Sourcing from GAPartNet, Drag-a-Move~\cite{li2024dragapart} provides a 2D synthetic dataset for the task of part-oriented drag control. 
It contains a collection of triplets $(x, y, D)$, where image $x$ and $y$ are the object images in an initial and new states, and $D$ is a collection of drag actions applied to image $x$.
The drag annotation is projected from the ground-truth articulation annotations from 3D models to 2D images.
This dataset facilitates research on interactive generative models for articulated objects using drag control on 2D images with 3D awareness.

 Recently, S2O~\cite{iliash2024s2o} points out that PartNet-Mobility dataset suffers from lack of diversity, contains highly similar objects, and is biased towards objects with simple articulation structures.
So they propose a new dataset, Articulated Containers Dataset (ACD), which provides more challenging and realistic 3D objects.
The shapes are sourced from ABO~\cite{collins2022abo}, 3D-Future~\cite{fu20213dfuture}, and HSSD~\cite{khanna2024habitat} datasets.
The objects in ACD are more complex in geometry (e.g., with L-shaped tables and corner cabinets) and articulation structures with significantly more openable parts.
This dataset can potentially help to reduce the gap of models trained on synthetic data generalized to real objects.
\jy{On the track to increase the data diversity, Arti-PG~\cite{sun2024artipg} introduces a procedural generation toolbox that uses mathematical rules to synthesize articulated objects, with code as the user interface.
By randomizing configurations, this toolbox enables unlimited data variations, making it useful for scaling dataset volume.}

\subsection{Real-world Data}
A primary challenge in working with synthetic data is the unrealistic assumption that it is perfect—overly simplistic, complete, and free from noise.
 This can result in models that perform well in synthetic environments but struggle when applied to real-world scenarios, where data is often noisy and incomplete
To fill this synthetic-real gap, more recent efforts have started to focus on collecting data from real-world sensors, such as RGB-D cameras.

RBO~\cite{martin2019rbo} represents the pioneering dataset in this regard, being the first to collect data from RGB-D scans. 
It not only provides RGB-D recordings of human interaction with objects under varying experimental conditions, but also creates corresponding synthetic mesh models for each object. 
ReArt-48~\cite{liu2022toward} introduced a slightly larger dataset that reconstructs meshes from RGB-D scans and provides the articulation annotations across five object categories.
Taking this further, AKB-48~\cite{liu2022akb} emerges as the first large-scale dataset for articulated objects based on real scans,
where each object is described in a knowledge graph.
To construct this dataset, a fast articulation knowledge modeling pipeline is presented, significantly reducing the cost and effort required for object modeling in the real world. 
This innovation enables the creation of 3D models on a scale comparable to PartNet-Mobility.
Additionally, AKB-48 also annotates the physical properties of the objects, such as mass, to enhance the dataset's applicability. 
This additional annotation is important for bridging the generalization gap between simulation and real-world applications.

In parallel, the MultiScan~\cite{mao2022multiscan} dataset marks a groundbreaking advancement as the first large-scale scene-level dataset that documents multiple states of articulated objects in indoor settings.
To compile this dataset, a scalable 3D environment acquisition pipeline is designed for processing raw RGB-D scans to produce 3D surface mesh reconstruction with texture and articulation annotations for each articulated object in the scene.
MultiScan is an invaluable resource for advancing research and applications that require an understanding of the dynamics and interactions of articulated objects in real-world environments at the scene level.
Leveraging the MultiScan dataset, OPDMulti~\cite{sun2023opdmulti} is introduced for the OPD task of the multiple-parts version by extracting frames from the RGB-D scans and annotating the openable parts in each frame.
It extends the OPD task to the scene level, which is more challenging and practical in real-life scenarios.

ParaHome~\cite{kim2024parahome} is a recent real-world dataset that captures 3D human motions and interactions with objects within a home environment.
The authors set up a system to capture the human motions with wearable motion capture devices and to track the dynamic interactions with the object synchronized with multi-view RGB cameras.
Articulated objects with multiple parts in the dataset are expressed with parameterized articulations.
This dataset enables new opportunities for human-object interaction studies, which is an important application in the field of embodied AI and robotics.

To address the task of articulation estimation from 2D, OPDReal~\cite{jiang2022opd} dataset is introduced by collecting RGB-D scans from real-world scenes.
By providing a set of real RGB images capturing various articulation states of the objects, along with reconstructed mesh models, OPDReal presents new opportunities for research in the field of articulated object modeling from 2D visuals.
OPDMulti~\cite{sun2023opdmulti} follows this trend and extends the task to multiple objects in the scene by collecting RGB-D frames, which is more challenging and practical in real-life scenarios.
As a subtask that tackles a segmentation problem for multiple articulated parts, \citet{wang2025active} introduces a 2D dataset that consists of 2,550 RGB images captured fromthe  real world.
This dataset focuses on articulated part segmentation by leveraging an active learning model to get fine-grained annotation for part segmentation, saving the cost of manual annotation.
However, this dataset does not provide articulation annotations for each part.

\input{tabs/geometry.tex}

%% file: tabs/geometry.tex
\begin{table*}
    \centering
    \resizebox{\textwidth}{!}{
    \begin{tabular}{@{}lllccccllllc@{}}
    \toprule
    & \multicolumn{2}{c}{Input} & \multicolumn{4}{c}{Input Assumptions} & \multicolumn{3}{c}{Methodology} & \multicolumn{2}{c}{Output}   \\ 
    \cmidrule(l){2-3}\cmidrule(l){4-7}\cmidrule(l){8-10}\cmidrule(l){11-12}
                                                & geo rep.      & \# states & \# parts  & partial/noisy & aligned   & \jy{part seg.}    & intermediate rep. & strategy        & supervision   & geo rep.    & arti. part seg.  \\ \midrule
    \multicolumn{12}{l}{Articulated Part Perception} \\ \midrule 
    \citet{xu2009joint}                         & mesh          & 1         & \xmark    & \xmark        & -         & \jy{ \xmark }           & surface patch     & handcrafted     & -             & mesh        & \cmark     \\
    \citet{mitra2010illustrating}               & mesh          & 1         & \xmark    & \xmark        & -         & \jy{ \cmark }           & surface patch     & handcrafted     & -             & mesh        & \xmark     \\
    \citet{sharf2014mobility}                   & mesh          & multi     & \xmark    & \cmark        & \xmark    & \jy{ \xmark }           & surface patch     & handcrafted     & -             & mesh        & \cmark     \\
    \citet{yuan2016space}                       & PC            & multi     & \xmark    & \cmark        & \xmark    & \jy{ \xmark }           & 3D trajectory     & handcrafted     & -             & PC          & \cmark     \\ 
    \citet{li2016mobility}                      & RGB-D         & multi     & \xmark    & \cmark        & \xmark    & \jy{ \xmark }           & 3D trajectory     & handcrafted     & -             & PC          & \cmark     \\   
    \citet{hu2017learning}                      & mesh          & 1         & \cmark    & \xmark        & -         & \jy{ \cmark }           & surface patch     & -               & -             & mesh        & \xmark     \\
    \citet{yi2018deep}                          & PC            & 2         & \xmark    & \cmark        & \xmark    & \jy{ \xmark }           & PC feat.          & SL              & labeled PC    & PC          & \cmark     \\    
    Shape2Motion~\cite{wang2019shape2motion}    & PC            & 1         & \xmark    & \xmark        & -         & \jy{ \xmark }           & PC feat.          & SL              & labeled PC    & PC          & \cmark     \\
    RPM-Net~\cite{yan2019rpm}                   & PC            & 1         & \xmark    & \cmark        & -         & \jy{ \xmark }           & PC feat.          & SL              & labeled PC    & PC          & \cmark     \\
    \citet{abbatematteo2019learning}            & RGB-D         & 1         & \cmark    & \cmark        & -         & \jy{ \xmark }           & image feat.       & SL              & prim. params. & prim. params.  & \cmark     \\
    ANCSH~\cite{li2020category}                 & PC            & 1         & \cmark    & \cmark        & -         & \jy{ \xmark }           & PC feat.          & SL              & labeled PC    & PC          & \cmark     \\
    \citet{shi2021self}                         & PC            & multi     & \xmark    & \cmark        & \cmark    & \jy{ \xmark }           & PC feat.          & SSL             & PC            & PC          & \cmark     \\
    MultiBodySync~\cite{huang2021multibodysync} & PC            & multi     & \xmark    & \cmark        & \cmark    & \jy{ \xmark }           & PC feat.          & SSL             & PC            & PC          & \cmark     \\
    CAPTRA~\cite{weng2021captra}                & PC            & multi     & \cmark    & \cmark        & \cmark    & \jy{ \xmark }           & PC feat.          & SL              & labeled PC    & PC          & \cmark     \\
    \citet{abdul2022learning}                   & PC            & 1         & \xmark    & \cmark        & -         & \jy{ \xmark }           & PC feat.          & SL              & labeled PC    & PC          & \cmark     \\
    \citet{qian2022understanding}               & RGB           & multi     & \xmark    & \cmark        & \cmark    & \jy{ \xmark }           & image feat.       & SL              & labeled PC    & 3D plane    & \cmark     \\
    OPD~\cite{jiang2022opd}                     & RGB           & 1         & \cmark    & -             & -         & \jy{ \xmark }           & image feat.       & SL              & 2D mask       & 2D mask     & \cmark     \\
    OPDMulti~\cite{sun2023opdmulti}             & RGB           & 1         & \xmark    & -             & -         & \jy{ \xmark }           & image feat.       & SL              & 2D mask       & 2D mask     & \cmark     \\
    \citet{liu2022toward}                       & RGB-D         & 1         & \xmark    & \cmark        & -         & \jy{ \xmark }           & PC feat.          & SL              & labeled PC    & PC          & \cmark     \\
    GAPartNet~\cite{geng2023gapartnet}          & RGB-D         & 1         & \cmark    & \cmark        & -         & \jy{ \xmark }           & PC feat.          & SL              & labeled PC    & PC          & \cmark     \\
    \citet{liu2023building}                     & PC            & multi     & \xmark    & \cmark        & -         & \jy{ \xmark }           & PC feat.          & SSL             & PC            & PC          & \cmark     \\
    \citet{liu2023self}                         & PC            & 1         & \cmark    & \cmark        & -         & \jy{ \xmark }           & PC feat.          & SSL             & PC            & PC          & \cmark     \\
    \citet{liu2023semi}                         & PC            & 1         & \xmark    & \xmark        & -         & \jy{ \cmark}            & PC feat.          & \jy{semi-weakly SL}  & labeled PC    & PC     & \cmark     \\
    Banana~\cite{deng2023banana}                & PC            & 1         & \cmark    & \xmark        & -         & \jy{ \xmark }           & PC feat.          & SL              & labeled PC    & PC          & \cmark     \\
    \citet{wang2025active}                      & RGB           & 1         & \xmark    & \cmark        & -         & \jy{ \xmark }           & image feat.       & SL              & 2D mask       & 2D mask     & \cmark     \\
    \jy{AutoURDF~\cite{lin2024autourdf}}        & PC            & multi     & \xmark    & \cmark        & \cmark    & \jy{ \xmark }           & neural field      & SSL              & PC           & PC          & \cmark     \\
    GAMMA~\cite{yu2024gamma}                    & PC            & 1         & \xmark    & \cmark        & -         & \jy{ \xmark }           & PC feat.          & SL              & labeled PC    & PC          & \cmark     \\

    \midrule \multicolumn{12}{l}{Articulated Object Creation} \\ \midrule
    \citet{pekelny2008articulated}              & PC            & multi     & \cmark    & \cmark        & \cmark    & \jy{ \cmark}            & PC                & handcrafted     & -             & mesh        & \cmark     \\
    A-SDF~\cite{mu2021asdf}                     & SDF           & 1         & \cmark    & \xmark        & \cmark    & \jy{ \xmark}            & neural field      & SL              & SDF           & mesh        & \xmark     \\
    Ditto~\cite{jiang2022ditto}                 & PC            & 2         & \cmark    & \cmark        & \cmark    & \jy{ \xmark}            & neural field      & SL              & labeled PC    & mesh        & \cmark     \\
    CLA-NeRF~\cite{tseng2022cla}                & MV RGBs       & 1         & \cmark    & \cmark        & \cmark    & \jy{ \xmark}            & neural field      & SL              & 2D mask       & mesh        & \cmark     \\
    \citet{wei2022self}                         & MV RGBs       & 1         & \cmark    & \cmark        & \cmark    & \jy{ \xmark}            & neural field      & SL              & MV RGBs       & mesh        & \xmark     \\
    \jy{WatchItMove~\cite{noguchi2022watch}}    & MV RGBs       & multi     & \xmark    & \xmark        & \cmark    & \jy{ \xmark}            & neural field      & SSL             & MV RGBs       & ellipsoids  & \cmark     \\
    CARTO~\cite{heppert2023carto}               & MV RGBs       & 1         & \cmark    & \cmark        & -         & \jy{ \xmark}            & neural field      & SL              & SDF           & mesh        & \xmark     \\
    \citet{liu2023few}                          & -             & -         & \cmark    & -             & \cmark    & \jy{ \cmark}            & convex            & TL              & -             & mesh        & \cmark     \\
    PARIS~\cite{liu2023paris}                   & MV RGBs       & 2         & \cmark    & \cmark        & \cmark    & \jy{ \xmark}            & neural field      & SSL             & MV RGBs       & mesh        & \cmark     \\
    SfA~\cite{nie2023sfa}                       & RGB PC        & multi     & \xmark    & \cmark        & \cmark    & \jy{ \xmark}            & neural field      & SL              & PC            & mesh        & \cmark     \\
    NAP~\cite{lei2023nap}                       & -             & -         & \xmark    & -             & \cmark    & \jy{ \xmark}            & latent feat.      & generative      & -             & mesh        & \cmark     \\
    CAGE~\cite{liu2024cage}                     & -             & -         & \xmark    & -             & \cmark    & \jy{ \xmark}            & latent feat.      & generative      & -             & mesh        & \cmark     \\
    \citet{weng2024neural}                      & MV RGB-Ds     & 2         & \cmark    & \cmark        & \cmark    & \jy{ \xmark}            & neural field      & SSL             & MV RGB-Ds     & mesh        & \cmark     \\
    REACTO~\cite{song2024reacto}                & MV RGBs       & multi     & \cmark    & \cmark        & \cmark    & \jy{ \xmark}            & neural field      & SSL             & MV RGBs       & mesh        & \xmark     \\
    Real2Code~\cite{mandi2024real2code}         & RGBs          & 1         & \xmark    & \cmark        & -         & \jy{ \xmark}            & PC, neural field  & \jy{TL}         & occupancy     & mesh        & \cmark     \\
    RSRD~\cite{kerr2024robot}                   & MV RGBs       & multi     & \xmark    & \cmark        & \cmark    & \jy{ \xmark}            & neural field      & SSL             & MV RGBs       & mesh        & \cmark     \\
    DragAPart~\cite{li2024dragapart}            & RGB           & 1         & \xmark    & \cmark        & -         & \jy{ \xmark}            & image feat.       & generative      & RGB           & RGB         & \xmark     \\
    LEIA~\cite{swaminathan2024leia}             & MV RGBs       & 4         & \xmark    & \cmark        & \cmark    & \jy{ \xmark}            & neural field      & SSL             & MV RGBs       & mesh        & \cmark     \\
    \jy{PhysPart~\cite{luo2024physpart}}        & -             & -         & \xmark    & -             & \cmark    & \jy{ \xmark}            & latent feat.      & generative      & -             & mesh        & \cmark     \\
    \jy{SINGAPO~\cite{liu2024singapo}}               & -        & -         & \xmark    & -             & \cmark    & \jy{ \xmark}            & latent feat.      & generative      & -             & mesh        & \cmark     \\
    \jy{Articulate-Anything~\cite{le2024articulate}} & -        & -         & \xmark    & -             & \cmark    & \jy{ \xmark}            & latent feat.      & \jy{TL}         & -             & mesh        & \cmark     \\
    \jy{MeshArt~\cite{gao2024meshart}}               & -        & -         & \xmark    & -             & \cmark    & \jy{ \xmark}            & latent feat.      & generative      & -             & mesh        & \cmark     \\
    \jy{ArtFormer~\cite{su2024artformer}}            & -        & -         & \xmark    & -             & \cmark    & \jy{ \xmark}            & latent feat.      & generative      & -             & mesh        & \cmark     \\
    \jy{Articulate AnyMesh~\cite{qiu2025anymesh}}   & mesh      & 1         & \xmark    & \xmark        & \xmark    & \jy{ \xmark}            & MV RGBs           & \jy{TL}         & -             & mesh        & \cmark     \\
    ArtGS~\cite{liu2025building}                    & MV RGB-Ds      & 2         & \xmark    & \xmark        & \cmark    & \jy{ \xmark}            & neural field      & SSL             & MV RGB-Ds       & mesh        & \cmark     \\
    \bottomrule
    \end{tabular}
    }
    \caption{Summary of the work in  articulated object modeling from the geometric perspective.
    The table provides information about the geometric representation and the number of articulted states observed in the input, the assumptions made on the input data, the methodology used in the geometry  modeling, and the output data representation.
    }
    \label{tab:geo}
\end{table*}

%% file: secs/4_geometric.tex
\section{Geometry Modeling}
\label{sec:geometric}
In this section, we discuss the recent works on articulated object modeling from the geometry modeling perspective.
Based on the goal of the tasks, we categorize the works into two groups: \textit{articulated part perception} and \textit{articulated object creation}.
The former group of work focuses on understanding the articulated structure of the objects from 3D or 2D observations in terms of the articulated part analysis and articulation parameters estimation, and in this section, we focus the discussion on the subtask of shape analysis in each work.
The latter group focuses on reconstructing or generating the 3D shape of objects either with motion parameters associated for each part or with the object animated into different articulation states over time.
In \Cref{tab:geo}, we summarize these works in terms of the choice of geometric representation, the assumptions that each work made for the input, and the methodology each work adopted for the articulated shape analysis.
We specify the meaning of each column in \Cref{tab:geo} as follows:
\begin{itemize}
    \item \textbf{geo rep.}: the geometric representation as input, output, or intermediate representation during processing (under \textit{methodology}).
    \item \textbf{\# states}: the number of articulation states observed in the input.
    \item \textbf{\# parts}:  as an assumption, whether the number of articulated parts is known in the input object.
    \item \textbf{partial/noisy}: whether the input geometry can be partially observed (e.g. point cloud projected from the single view depth or even without depth) or noisy (e.g. raw scans with sensory noise).
    \item \textbf{aligned}:  as an assumption, whether the object is pre-aligned across different states when multiple articulated states are observed in the input; whether the objects are aligned in a canonical space when there are multiple objects in the input;  whether the objects are assumed to be canonicalized for generative models.
    \item \textbf{part seg.}: \jy{whether the methods rely on precomputed part segmentation as the input assumption}.
    \item \textbf{intermediate rep.}: the intermediate representation used during the processing of the input shape or to represent the shape.  Please refer to \Cref{subsec:interm_rep} and \Cref{fig:interm_rep} for more explanation.
    \item \textbf{strategy}: the methodology adopted for shape analysis, including handcrafted methods, supervised learning (SL), \jy{semi- and weakly supervised learning (semi-weakly SL), transfer learning (TL)}, and self-supervised learning (SSL).
    \textit{Handcrafted} refers to non-learning-based methods that rely on human-designed descriptors, heuristic rules, or mathematical algorithms to analyze the shape. 
    \textit{SL} refers to the data-driven methods that require ground truth labeled data for training, such as segmentation labels, displacement for each point, etc.
    \textit{SSL} refers to the data-driven methods that only rely on the input data for training, such as leveraging the geometric consistency between different states of the object.
    \jy{\textit{Semi-weakly SL} refers to a hybrid learning paradigm that combines elements of semi-supervised learning and weakly supervised learning to leverage both limited strongly labeled data, abundant weakly labeled data, and unlabeled data for model training.
    \textit{TL} refers to the data-driven methods that adapt pre-trained foundational models for a specific target task or use a pre-trained model as a feature extractor to adapt to a related task.}
    \item \textbf{supervision}: the source of supervision used for the learning-based methods, such as segmentation labels on the point cloud (\textit{labeled PC}), segmentation mask on images (\textit{2D mask}),  signed distance field (\textit{SDF}) or occupancy field (\textit{occupancy}) of the object, multi-view RGB(-D) images paired with camera parameters (\textit{MV RGBs} or \textit{MV RGB-Ds}), and unposed RGB images (\textit{RGBs}).
    \item \textbf {arti. part seg.}: the shape of the object is segmented into articulated parts as the output.
\end{itemize}

\subsection{Geometric Representation for Task Setting}
\subsubsection{Articulated Part Perception}
\label{subsec:geo_rep_percept}
\jy{
To analyze the part mobility of articulated objects, segmenting articulated parts from an input shape is typically a necessary step.
Exceptions are the early work by \citet{mitra2010illustrating} and \citet{hu2017learning} which simplified the problem by assuming articulated part segmentation was provided as input, 
and ScrewNet~\cite{jain2021screwnet} bypassing explicit segmentation by directly analyzing mobility from a point cloud sequence.
Beyond these, other articulated part perception methods jointly address the articulated part segmentation and mobility analysis from unstructured input shapes, with the exception of \citet{liu2023semi} which first leverages fine-grained semantic segmentation to group articulated parts for motion analysis and later uses them as weak labels to enhance perception performance.
In the following, we categorize existing works based on the geometric representation they take as input and discuss the assumptions made about the input data.
}

\input{figures/4_mobtree_sharf.tex}

\mypara{Perception from meshes.}
As the most widely used geometric representation in the industry, polygonal meshes are naturally chosen by early works~\cite{xu2009joint,mitra2010illustrating,sharf2014mobility,hu2017learning} as the initial input representation to work with.
These meshes can be obtained from CAD models that are carefully designed by the artists with part-level structure and sharp edges.
It implies that the input geometry in these works is assumed to be complete and noise-free, which simplifies the problem by observing the object in a geometrically perfect condition in 3D.
\citet{xu2009joint} takes a complex mesh model as input and segments the model into components for joint analysis used for articulate or deform the model.
The proposed method is demonstrated on a range of models, including connected multi-component models, disconnected multi-component models, and single-component models.
\citet{mitra2010illustrating} focus on the motion illustration for mechanical assemblies represented as a set of polygonal meshes that has been partitioned into individual parts.
They assume that each part is modeled as a 2-manifold surface with no self-intersections beyond a small tolerance.
\citet{sharf2014mobility} pioneered addressing the mobility analysis for indoor scenes represented as unorganized boundary meshes.
They did not make any assumptions on the mesh properties, such as self-intersection or non-manifoldness. 
However, segmenting objects and their parts from a scene requires observing state variations in the object and its components, as illustrated in \Cref{fig:mobtree_sharf}.
\citet{hu2017learning} learns part mobility model from a single snapshot of the object represented as polygonal meshes.
It assumes that the part segmentation is given as input and the method applies on one pair of reference-moving parts at a time.
\jy{A recent work by \cite{qiu2025anymesh} proposes to convert a static mesh model into its articulated counterparts in a open-vocabulary manner.
It does not assume any pre-computed part segementation and category information given as input.
This approach leverages the recent advances in image segmentation models to assist the 3D segmentation and use VLMs for the part mobility analysis.}

\jy{
The main goal of these mesh-based works is to analyze the mobility on artist-created models or manufacturing designs, which can be potentially useful in automating the animation process, illustrating the motion of mechanical assemblies, or creating interactive simulations.
Some challenges are commonly shared among these works, such as the complexity in processing varying mesh structures with irregular topology and the difficulty in converting the mesh into a more compact representation for learning-based methods.
The assumption of having clean and complete geometry in the input also limits the applicability of these methods to real-world object reconstructions, which are often incomplete and noisy.
}

\mypara{Perception from point clouds.}
Point clouds of the objects can be obtained from 3D scans or depth sensors, which are more accessible and easier to obtain than the polygonal meshes.
This accessibility makes the point clouds a more practical choice for the input representation as it can be directly obtained from the real world.
It encourages more works later that consume point clouds as input for articulated part perception  (examples shown in \Cref{fig:pcinput}).
\jy{Leveraging this ease of acquisition, a line of work~\cite{li2016mobility,yuan2016space,shi2021self,shi2021self,jain2021screwnet,weng2021captra,nie2023sfa,liu2023building} takes point clouds that observe a temporal sequence of articulated motion of the object as input.
MultiBodySync~\cite{huang2021multibodysync} also processes multiple scans but does not assume a temporal sequence.
These point cloud sequences are often incomplete and noisy, which makes the robustness of the motion fitting an additional critical requirement for the system.
Another implied assumption is that the object is aligned in the same coordinate system across different states, which is a prerequisite for the motion fitting process.
By registering the point positions over time, the points exhibiting consistent motion patterns are grouped as the articulated parts.
The motion patterns can be further analyzed to estimate the articulation parameters for each part.
However, observing a sequence showing object motion can sometimes be unnatural for objects that cannot move on their own.
Consequently, the motion-capturing process typically demands extensive human intervention, making it costly and labor-intensive.
Another challenge in the motion capturing process is how to filter out the irrelevant components in the observed sequence, such as the human hand or robotic arm.
}

Instead of observing motion sequences, \citet{yi2018deep} and Ditto~\cite{jiang2022ditto} take a pair of point clouds of different articulation states as input.
\citet{yi2018deep} relaxes several assumptions made by the previous works that the object can be misaligned in two states and the input only needs to be from objects that are functionally related but not necessarily from the same instance.
Ditto assumes the point cloud pair is pre-aligned from the same object and also assumes only one part is moving in the observation.
But it produces the mesh surface with each part segmented as the output to build digital twins of the object.
More recent works~\cite{wang2019shape2motion,yan2019rpm,li2020category,liu2023self,deng2023banana} tackle the problem from a single state of the object at inference time to make the input more accessible.
\citet{abdul2022learning}, \citet{liu2022toward}, and GAPartNet~\cite{geng2023gapartnet} further extend the input to colored point clouds, which can leverage more visual information than the raw point clouds.
\jy{In tackling the challenge of lacking training data with ground-truth labels, \citet{liu2023semi} proposes to leverage data with fine-grained semantic segmentation from PartNet~\cite{mo2019partnet} as weak labels to augment the training data for the part mobility analysis.}

\jy{
Perceiving part mobility from point clouds facilitates the analysis of real-world objects with reduced domain transfer effort compared to mesh-based methods.
Point clouds are easier to obtain from depth sensors, making them a practical choice for many applications. 
As a discrete approximation of object surfaces, they are flexible, lightweight, and easier to process with learning-based methods. 
However, their lack of connectivity information and fine surface details makes them less informative than meshes, which can capture intricate geometric features more effectively. 
This limitation poses a challenge for learning-based methods, as they must extract meaningful features from sparse and unstructured data to accurately understand part structures.
}

\mypara{Perception from monocular RGB images.}
Even more readily available than point clouds, RGB(-D) images are another common input for articulated part perception.
\citet{qian2022understanding} proposed to extract 3D planes to represent articulated parts from RGB videos recording the human-object interaction.
OPD~\cite{jiang2022opd} first introduces the task of detecting openable parts of objects from a single-view image.
OPDMulti~\cite{sun2023opdmulti} extends the scenario to multiple objects in the scene as a follow-up.
\citet{wang2025active} further improves the accuracy of the articulated part segmentation from real images by leveraging active learning to easily obtain the ground truth labels and propose a new dataset for the task.
\jy{Compared to the aforementioned methods that rely solely on 3D geometry, RGB-based approaches eliminate the need for extensive 3D capturing or reconstruction, making them highly beneficial for real-time applications. 
These methods leverage visual cues in images, providing richer appearance details and contextual information about the object. 
However, they also face unique challenges, including the difficulty of inferring 3D structures from 2D images, handling occlusions and scene clutter, and compensating for limited or missing 3D information in the input.
}

\subsubsection{Articulated Object Creation}
In recent years, significant advances in neural representation techniques~\cite{mildenhall2021nerf,wang2021neus,kerbl20233d}, along with large vision models~\cite{oquab2023dinov2,kirillov2023sam,wang2024dust3r} and generative models~\cite{goodfellow2014gan,ho2020ddpm,rombach2022high,blattmann2023stable}, have opened new possibilities for 3D modeling for articulated objects. 
These approaches enable more flexible and detailed representations of articulated objects, which have traditionally posed challenges due to their complex geometries and dynamic characteristics. 
This emerging direction in articulated object creation leverages the strengths of both neural geometry representations and generative models, facilitating innovative methods for reconstructing or generating articulated objects across various scenarios.

\mypara{Articulated object reconstruction.}
The goal of articulated object reconstruction is to recover the 3D geometry and motion parameters of the objects from various observations.
On the geometry side, the challenges not only involve decomposing the part into articulated components in the perceived format, but also producing the reconstructed surface that can be animated into different articulation states.
The input modalities can be diverse, including point clouds, multi-view RGB or RGB-D images, and RGB videos.
Certain assumptions are made about the input data based on the target object types and the sparsity of observations. 
These include factors such as the number of articulated parts, the object's alignment across different states, and the completeness of the input data.
The output representation is typically the mesh surface, which is widely used for animation and simulation.
A line of work focuses on reconstructing the whole object as a single deformable surface either over time or given different states~\cite{mu2021asdf,wei2022self,heppert2023carto,song2024reacto,swaminathan2024leia,kerr2024robot}.
Another line of work reconstructs each part as separate surfaces interconnected by joints to enable articulation~\cite{tseng2022cla,jiang2022ditto,liu2023paris,weng2024neural,mandi2024real2code}.

\input{figures/4_arti_gen.tex}

\citet{pekelny2008articulated} pioneered surface reconstruction from sequential point clouds, assuming known part segmentation and skeleton in the first frame.
\jy{Building on this idea of observing objects in motion, WatchItMove~\cite{noguchi2022watch} extends the input to multi-view RGB videos and approximates each moving part as an ellipsoid.
RSRD~\cite{kerr2024robot} models objects in 4D space by taking multi-view RGB images of their static state and a casual RGB video demonstrating human interaction.}
Similarly, REACTO~\cite{song2024reacto} proposes to reconstruct 4D objects from a single causally captured monocular RGB video of the object in motion.
While these methods accommodate arbitrary kinematic structures by registering different object states, they require dense motion observations.
A-SDF~\cite{mu2021asdf} was the first to reconstruct articulated objects using signed distance fields from a single snapshot, allowing deformation into different articulation states.
\citet{wei2022self} share the same goal with A-SDF but take multi-view RGB images as input for the reconstruction, while CARTO~\cite{heppert2023carto} further extends the input to a single stereo RGB image.
These works ease the data acquisition process by requiring only a single snapshot of the object, but they assume the object has only one movable part.
LEIA~\cite{swaminathan2024leia} reconstructs the interpolated state of the object between the two given input states from multi-view images without knowing the number of moving parts.
\jy{However, one limitation of the above-mentioned works is that they all reconstruct the articulated object as a single unified surface or as an implicit deformation field.
The lack of part-level structure in the output geometry makes it difficult to interact with the object in a physically realistic manner.
}

Recent works address this limitation by reconstructing the articulated object in the part level.
SfA~\cite{nie2023sfa} designs a pipeline to reconstruct real-world objects by iteratively interacting with the moving parts and capture the object in the point cloud sequence.
To make the reconstruction more efficient, CLA-NeRF~\cite{tseng2022cla} proposes to reconstruct the object from a few multi-view RGB images of an unseen 3D object instance within the known category.
It assumes that the object category is known and all the instances in the category share the same articulated structure.
To avoid the reliance on object category, Ditto~\cite{jiang2022ditto} reconstructs from a pair of point clouds, while PARIS~\cite{liu2023paris} take multi-view RGB images observing the object in two states as input.
\citet{weng2024neural} extends the setting of PARIS to multi-view RGB-D images and is able to reconstruct multiple articulated parts in the output.
\jy{These works can reconstruct the object across arbitrary categories, but they all assume the number of articulated parts is known at inference time.
In a more flexible setting, Real2Code~\cite{mandi2024real2code} proposes to reconstruct articulated parts from multiple unposed RGB images of the object in an opening state.
This work focuses on certain categories of objects that mostly have cuboid-like parts but with no assumption on the number of articulated parts.
Capturing the opening state of the object is also helpful in modeling the interior geometry of the object, which is usually occluded in the closed state.
However, this method relies on the object being observed in a state that maximally exposes each part, which may be constrained by its natural configuration or the observation angle.
}

\jy{
Despite the rapid progress in articulated object reconstruction, existing methods still face several challenges.
Early approaches primarily focused on surface reconstruction from sequential point clouds or dense motion observations but lacked explicit part-level structure, making realistic interaction difficult. 
Recent works have addressed this by reconstructing objects at the part level, but they often rely on strong priors such as known object categories, predefined articulation structures, or a fixed number of moving parts.
This highlights a critical gap in the field: the need for methods that can \emph{reconstruct arbitrary articulated objects with unknown part structure from sparse and unconstrained observations}.
}

\mypara{Articulated object generation.}
This line of work aims to synthesize the articulated objects in both geometric and kinematic structures from random noise or user-specified constraints.
The representative work is illustrated in \Cref{fig:arti_gen}.
This is a newly emerging direction of research that leverages the recent advances in neural representation learning and generative modeling in creating high-fidelity and diverse 3D assets.
These generative models learn the underlying distribution of articulated objects, capturing articulated part geometry and articulation configurations to generate novel samples.
NAP~\cite{lei2023nap} is a pioneering work that unconditionally generates a full description of an articulated object, including part geometry, articulation graph, and joint parameters.
\jy{MeshArt~\cite{gao2024meshart} tackles the same task but designs a different geometry representation by directly generating the mesh vertices from the latent code.
CAGE~\cite{liu2024cage}, SINGAPO~\cite{liu2024singapo}, and ArtFormer~\cite{su2024artformer} introduce different controllable generation setting:
CAGE incorporates high-level structural information (e.g., articulation graph) as user constraints, 
SINGAPO enables single-image-driven generation, 
and ArtFormer focuses on text-driven generation.
Instead of learning the distribution from scratch, ARTICULATE-ANYTHING~\cite{le2024articulate} proposes a general framework that automates the articulation of objects from different input modalities, including text, images, and videos.
Programs are used to arrange the part geometries and predict articulation with a critic mechanism designed to ensure the plausibility of the generated object.
The output of these works is a mesh representation that is either synthesized from the intermediate representation or retrieved from a database using the object context.
A key challenge in this direction is representing articulated objects in a \emph{compact and expressive way to capture the joint distribution of part geometry and articulation} effectively.
Another challenge is enabling \emph{precise control over the generation process}, especially for complex, highly structured objects. 
}

On a different front, DragAPart~\cite{li2024dragapart} introduced a method to generate images of articulated objects in different states from a single input image, driven by user-specified drag actions. 
While the output remains in 2D space, its articulation is 3D-aware, \jy{allowing users to interactively explore the object's motion in a 2D setting.
This highlights an under-explored yet promising direction: simulating 3D-aware articulation dynamics within a 2D framework.}

\subsection{Methodology}

\subsubsection{Intermediate Representation}
\label{subsec:interm_rep}
In this section, we discuss the different choices of intermediate geometric representations used for part mobility analysis or object reconstruction, some of which are visualized in \Cref{fig:interm_rep}.

\mypara{Articulated part perception.}
 In the task of articulated part perception, various intermediate representations are employed to support geometric analysis, either through neural networks or handcrafted methods. 
These intermediate representations serve as critical abstractions that simplify the complex structures of articulated objects, making them more tractable for further processing. 
Depending on the type of input data, these representations can take different forms, including surface patches for capturing local geometric details, 3D trajectories for tracking motion over time, point cloud features for representing spatial relationships, or image features for extracting visual cues from 2D data. 
Each representation is chosen based on the specific demands of the task, and it plays a crucial role in bridging the gap between raw perception data and high-level geometric reasoning.

\begin{itemize}
    \item \textbf{Surface patch}. The early methods taking mesh as input commonly leverage the surface patches as the proxy for analysis.
    The local patches can be detected from slippable analysis as kinematic surfaces~\cite{gelfand2004shape}, which are used to associate with rigid motions for each part~\cite{xu2009joint}.
    The surface patches can also be useful for revealing the self-similarity and symmetry on the mechanical part to characterize the motion configuration~\cite{mitra2010illustrating}.
    The patches at the intersection area between components can be leveraged to determine the supporting-supported relationship~\cite{sharf2014mobility}, which are informative for the part decomposition and structure analysis.
    Similarly, the interaction pattern described at the connecting region between surface patches is leveraged in \citet{hu2017learning} to measure the similarity between snapshots of the object for motion classification.
    \item \textbf{3D trajectory}. The 3D trajectory is a spatial-temporal representation that records the position of the points in the 3D space over time.
    For early works capturing motion with a sequence of point clouds as input~\cite{pekelny2008articulated,yuan2016space}, explicitly extracting the 3D trajectory is a common choice as the intermediate representation to optimize the correspondences between points along the timeline. 
    Once the trajectory is robustly estimated, the rigid pieces in the object can be grouped based on the transformational consistency of the points.
    \item \textbf{Point cloud feature (PC feat.)}. Point cloud latent features are commonly used as the intermediate representation in the methods that take point clouds as input.
    The learning-based methods typically extract spatial point-wise features from the input point cloud and supervise them for the specific objectives, such as part labeling, point registration, or articulation parameter estimation.
    PointNet~\cite{qi2016pointnet}, PointNet++~\cite{qi2017pointnet++}, and Sparse U-Net~\cite{graham2018submanifold} are the common feature extractors used as a backbone method in existing works.
    For the works that takes point cloud sequence or pair as input~\cite{yi2018deep,shi2021self,huang2021multibodysync,weng2021captra,liu2023building,lin2024autourdf}, the point cloud features are used to capture point correspondence for temporal consistency across different states.
    For other works that take single state point cloud as input~\cite{wang2019shape2motion,yan2019rpm,li2020category,liu2023self,deng2023banana,abdul2022learning,liu2022toward,geng2023gapartnet,liu2023semi}, the point cloud features are used to encode the spatial relationships between points for articulated part labeling, while additional heads are usually added to predict the motion parameters or part pose along with the part segmentation.
    \item \textbf{Convex}. The convex is used to approximate the input geometry by representing the part as a convex hull.
    \citet{liu2023few} proposes to learn deformation networks by choosing convexes as intermediate representation to deform each part mesh to generate new objects by few-shot learning.
    \item \textbf{Image feature (image feat.)}. Image latent features are used as the intermediate representation in the works that take RGB or RGB-D images as input.
    Similar to the point cloud features, the image features are extracted from pixels for later use in mobility part analysis.
    ScrewNet~\cite{jain2021screwnet} follow this scheme to extract 2D features from depth images using ResNet~\cite{he2016deep} instead of per-point features from point clouds.
    For the work focused on articulation perception from a single image~\cite{jiang2022opd,sun2023opdmulti,wang2025active}, MaskRCNN~\cite{he2017mask} and Mask2Former~\cite{cheng2022masked} are the common backbones used to extract feature maps for detecting and masking the parts.
    \jy{\item \textbf{MV RGBs}. Articulate AnyMesh~\cite{qiu2025anymesh} segments the input mesh by using muli-view images as proxy.
    Once the articulated parts are segmented in the rendered multi-view images, the part labels are projected back to the mesh by voting from the corresponding pixels.
    It introduces a effective way to leverage the advanced image segmentation models to assist the 3D segmentation task.}
\end{itemize}

\mypara{Articulated object reconstruction.}
In the context of articulated object reconstruction, neural fields have emerged as the predominant intermediate representation.
A neural field is an implicit function parameterized by a neural network, enabling the continuous and differentiable representation of an object's 3D geometry, with the optional capability to model its appearance.
Once constructed, the neural field can be queried to reconstruct the objects and further converted into a mesh representation using the marching cube algorithm~\cite{lorensen1998marching}.
Depending on its formulation, existing works represent the object or their parts with signed distance functions (SDFs), occupancy fields, neural radiance fields (NeRFs), or 3D Gaussians.
\begin{itemize}
    \item \textbf{Neural SDF}. SDFs define the geometry of an object by encoding the signed distance from any point in space to the object's surface, with positive values outside the object and negative values inside.
    A-SDF~\cite{mu2021asdf}, \citet{wei2022self}, CARTO~\cite{heppert2023carto} are the works that leverage neural SDF to represent the object as a singular surface.
    The compactness of SDFs makes them effective at capturing smooth surfaces and can be trained to be deformable to animate the object into different articulation states.
    \jy{However, it becomes challenging to represent multiple parts with discrete surfaces in a single SDF, which limits the ability of this line of work to model articulated objects with multiple parts and complex structures.}
    \item \textbf{Neural occupancy field}. Occupancy fields are binary functions that indicate whether a point in space is occupied by the object or not.
    Ditto~\cite{jiang2022ditto} constructs a neural occupancy field from a pair of point clouds to represent the object with additional labels for each point in the field to indicate the part association.
    Real2Code~\cite{mandi2024real2code} also learns an occupancy-based shape completion network to model part surfaces.
    Both methods use occupancy fields to represent articulated objects from point cloud inputs.
    Occupancy fields provide flexible representations which effectively handle non-watertight and fragmented structures.
    \jy{However, like SDFs, occupancy fields also require dense 3D sampling as supervision to capture the surface details, leading to high training costs.}
    \item \textbf{NeRF}. NeRF is a continuous volumetric representation that models the object's geometry and appearance as a function of 3D coordinates.
    It is an effective representation that bridges the gap between 2D images and 3D representations via differentiable rendering.
    Several work uses NeRFs for articulated object reconstruct from multi-view RGB images~\cite{tseng2022cla,liu2023paris,swaminathan2024leia}, RGB-D images~\cite{weng2024neural} or RGB videos~\cite{song2024reacto}.
    The geometry of each part can be reconstructed by either learning separate NeRFs for each part~\cite{liu2023paris} or by associating part labels with the 3D coordinates~\cite{tseng2022cla,weng2024neural,song2024reacto}.
    LEIA~\cite{swaminathan2024leia} learns the interpolated state of the object between two given states by training a single NeRF.
    \jy{NeRFs excel at synthesizing realistic novel views of the objects under different articulated states once trained.
    But the quality of the extracted surface from NeRFs is easily affected by the sparsity of the training data, textureless regions, occlusions, and lighting conditions. 
    So they often require densely captured views for accurate reconstruction.} 
    \item \textbf{3D Gaussians}. 3D Gaussian representation models a scene or object using a set of anisotropic Gaussian functions in 3D space, each defined by position, shape, opacity, and color.
    3D Gaussian Splatting is a cutting-edge technique in neural rendering that enables efficient rendering, smooth interpolation, easy dynamic modeling, and high-quality reconstruction of the object.
    RSRD~\cite{kerr2024robot} first leverage this representation to model the dynamics of the articulated object over time as 4D Gaussian fields, which visually imitate the articulation from an object scan in multi-view and a monocular video.
    From the multi-view scan, they first reconstruct the static object using 3D Gaussians and cluster the Gaussians into semantic components.
    Each component is then tracked over time to capture the dynamics of the object parts by referring to the monocular video in DINO~\cite{caron2021emerging} feature space.
    ArtGS~\cite{liu2025building} leverages 3D Gaussians to reconstruct multi-part articulated objects given two states of the object in multi-view RGBD images.
    \jy{Compared to other more compact representations, such as SDFs, 3D Gaussians provide flexibility in representing multiple components by associating specific Gaussians with each part, enabling a more explicit way of capturing dynamics and interactions.
    3D Gaussian representation presents a promising step toward faster and more efficient reconstruction of articulated objects, which worth further exploration in the future.}
\end{itemize}

\input{figures/4_interm_rep.tex}

\jy{
The advantage of using implicit fields lies in their continuous and differentiable properties, making them well-suited for applications such as inverse rendering, shape fitting and completion, and deformation modeling. 
However, a major drawback is their computational expense, as each scene or object requires overfitting or optimization on a separate network, which limits their scalability and efficiency for large-scale or real-time use cases.
}
Real2Code~\cite{mandi2024real2code} bridges the 2D and 3D gap with depth estimation model~\cite{wang2024dust3r} and extract point cloud from the depth map.
The point cloud are then labeled as different parts by projecting the segmentation from multi-view images by leveraging SAM~\cite{kirillov2023sam}, which then completes the part surfaces by training a shape completion network.
This pipeline shows its advantage in computational efficiency and effectiveness by taking advantage of recent advances in foundational models.
But this approach relies on minimal occlusion in input images, otherwise the part geometry would be largely incomplete.

\mypara{Articulated object generation.}
In articulated object generation, designing an effective intermediate representation is crucial for capturing the complex structure and articulation of objects.
Given their inherently structured nature, articulated objects can be recorded in formats like URDF or MJCF for simulation in physics engines. 
Inspired by this, the intermediate representation for the generation task is designed to encode the part configuration in a hierarchical manner.
A common approach uses the part bounding box to anchor the spatial location of each part, while the surface geometry is compressed into a latent feature space if it is a part of the generation target.
See \Cref{fig:arti_gen} for a visualization of an example representation proposed in NAP~\cite{lei2023nap}.

NAP~\cite{lei2023nap} pioneers the design of the hierarchical representation as mentioned above.
To model the part surface, they trained an occupancy shape autoencoder~\cite{mescheder2019occupancy} to encode each part surface as a high-dimensional latent code which can be later decoded as SDFs.
CAGE~\cite{liu2024cage} \jy{and SINGAPO~\cite{liu2024singapo}} represent the parts with only bounding boxes with semantic labels as embeddings during the generation process, and then retrieve the part surface under the context of the generated part layout. 
This abstraction strategy allows the model to focus more on the structural arrangement and their interplay with the articulation at the generation stage.
\jy{ArtFormer~\cite{su2024artformer} further extends the representation from NAP and CAGE. 
It proposes to encode the part point cloud into triplane features and quantize the features into a discrete codebook, which is then used as latent code to generate the part geometry along with the structure.
The generated latent code can be decoded into SDFs to represent the part surface.
MeshArt follows the hierarchical representation design but tailors the geometry representation to directly generate the mesh vertices from the latent code.
As illustrated in ~\Cref{fig:meshart},
in the first generation stage, it represents each part's bounding box as a triangle mesh and learns a codebook as quantized triangle embeddings.
Guided by the global object structure generated in the first stage, another geometry codebook is learned for encoding the mesh triangles and their likelihood of being junction faces.
This design allows the model to directly generate triangle meshes for articulated parts, resulting in sharp and clean output surfaces.}

\jy{In contrast to the above works that model geometry and structure based on prior learning from training data distributions,
Articulate-Anything~\cite{le2024articulate} proposes to represent objects as programs that configure the part composition and articulation parameters.
An actor-critic system is developed to synthesize Python code that can be compiled into URDF files.
By rendering the object synthesized with the programs, the vision-language critic can provide feedback to refine the program generation.
This approach relies on the programmatic representation to ensure the plausibility of the generated object, which can be more interpretable and interactable than the learned latent space.
However, like CAGE and SINGAPO, it relies on a 3D assets library to retrieve the part geometry, which may limit the diversity of the generated objects.}

On the image-driven generation side, DragAPart~\cite{li2024dragapart} generates the part-level interaction in image space by encoding the drag-driven animation to the image features. 
They teach the pre-trained generative model to respond to drag prompt while enforcing physical consistency with respect to the underlying 3D geometry by fine-tuning the model with paired 2D images with drag annotations.
The motion prior learned by the model can be used to generate the object images in different states by dragging the part in the image.

\mypara{Discussion on intermediate representations.}
\jy{The choice of an intermediate representation depends on the input data modality, the application scenario, and the specific requirements of each task. 
For perception tasks, it is crucial to capture local geometric features and spatio-temporal relationships that enable part decomposition and motion analysis. 
Early work often relied on explicit forms (e.g., surface patches or 3D trajectories), while newer approaches employ implicit latent features from point clouds or images, reflecting a shift from handcrafted to learning-based techniques.
For creation tasks, representations must balance interpretability, computational demands, fidelity, and ease of real-world data acquisition. 
Implicit neural fields are popular for their continuous, differentiable nature—valuable in tasks like inverse rendering—but they typically require dense supervision and struggle to represent discrete part structures. 
A promising direction is to blend discrete and continuous representations (e.g., 3D Gaussians), which allows modeling piecewise-rigid parts in a differentiable manner.
The combination with the programmatic representation can further enhance the interpretability and controllability of the generated objects, which is beneficial for applications that require human interaction or physical simulation.
These hybrid representations also lend themselves to hierarchical structures, which is especially useful for modeling more structurally complex objects with multiple parts in a kinematic chain.
Along this direction, integrating graph representations or programs as domain-specific languages can be a promising approach to enhance the interpretability and controllability of generated objects with a hierarchical structure.
}

\subsubsection{Strategy and Supervision Signal}
\label{subsec:geo_strategy}
In this section, we discuss different algorithms and learning strategies adopted by the works analyzing the articulated part structure.

\mypara{Handcrafted methods.}
To infer part structure from the input geometry, early works rely on heuristic rules, human-designed descriptors, or geometric properties to analyze the shape.

For the works that take a single-state mesh as input, the approaches based on the \textit{slippage analysis}, \textit{geometric properties}, and \textit{relationship descriptors} are adopted to effectively segment the parts associated with the articulated motion.
The \textit{slippage analysis} is proposed by \citet{gelfand2004shape} to analyze the shape by discovering the slippable motions.
It can segment the geometry into slippable portions which they call kinematic surfaces to associate with rigid motions for each part.
Leveraging this approach, \citet{xu2009joint} detected primitive surfaces from the input object and link to motion joints that connect the articulated parts.
Geometric attributes on the mesh surfaces can provide valuable signals for inferring different joint configurations. 
Based on this insight, \citet{mitra2010illustrating} proposed to characterize the part that related to motions by detecting the geometrical primitives with \textit{self-similarity} and \textit{symmetry} on the mesh surface for each mechanical part. 
Spatial relationships between the parts are also informative for segmenting the components.
Building on this observation. \citet{sharf2014mobility} proposed to leverage the \textit{supporting-supported} relationship between surface patches to decompose a scene into a set of objects and their articulated parts.
Similarly, \citet{hu2017learning} proposed to leverage the \textit{interaction pattern} between parts described by a human-designed interaction descriptor~\cite{hu2015interaction}. 
Based on these geometric features, a distance function is designed to measure the similarity between snapshots of the object.

For the works that take point clouds observing a motion sequence as input,  geometric alignment algorithms, and handcrafted descriptors are the two common choices to find the correspondence between adjacent frames.
Once the correspondence is established,  the movement of each point over time can be tracked to construct the 3D trajectory.
Then the part segmentation can be inferred from the 3D trajectory by clustering the points that are moving together.
Given a known skeleton of the object in the first frame, \citet{pekelny2008articulated} proposed to trace the points that are skinned on each bone in the other frames using the \textit{Iterative Closest Point (ICP)} algorithm~\cite{besl1992method}.
Consuming an unlabelled point cloud sequence, \citet{yuan2016space} and \citet{shi2021self} proposed to leverage a deformable 3D shape registration algorithm~\cite{papazov2011deformable} to estimate the 3D trajectory.
Taking RGB-D sequence with additional color information as input, \citet{li2016mobility} proposed to use \textit{scene flow}~\cite{jaimez2015primal} to produce a raw dense correspondence, from which \textit{SIFT features}~\cite{lowe2004distinctive} are extracted from RGB images to refine and prune the initial proposals in the dense trajectory.

Handcrafted methods are excellent in terms of interpretability and computational efficiency, but their performance can be affected by the noiseness and incompleteness of the input data, initialization of the parameters in the algorithms, etc.
With the advent of machine learning and deep learning, handcrafted methods are gradually replaced by learning-based methods, which can automatically learn the underlying features from the data and adapt to the input conditions.
Later works after 2016 resort to data-driven methods to address the articulated object modeling problem, where the common practice converges into two main streams: supervised learning and self-supervised learning.

\begin{figure}[t]
    \begin{center}
    \includegraphics[width=\linewidth]{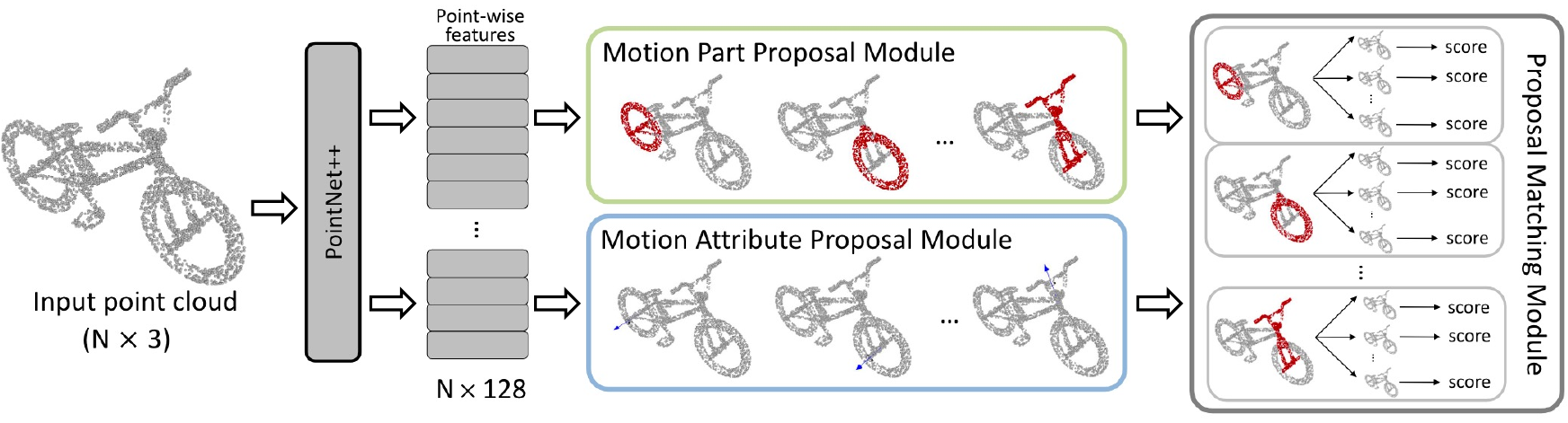}
    \end{center}
    \caption{Shape2Motion~\cite{wang2019shape2motion} jointly segments the articulated parts and estimates motion attributes from a single point cloud as input.
    Figure reproduced from original paper~\cite{wang2019shape2motion}.}
    \label{fig:shape2motion}
\end{figure}

\mypara{Supervised learning methods.}
On the perception side, a line of works proposes to learn the part decomposition from a single state of the object in a supervised manner.
\jy{A key distinction between articulated part segmentation and semantic part segmentation is that an articulated part may not be semantically atomic and can sometimes be composed of multiple semantic components. 
Unlike semantic segmentation, where labels can be consistently assigned across object instances, defining generalizable labels for articulated parts across categories is more challenging.
Shape2Motion~\cite{wang2019shape2motion} is the pioneering work that addresses the problem with a proposal-based solution, as illustrated in \Cref{fig:shape2motion}.
The model learns to regress a similarity matrix between points and a confidence score for each part proposal.
Using ground-truth part grouping, the similarity is supervised to minimize the point-wise feature distance among the points that belong to the same part, while the confidence map is trained to maximize the IoU between the point set of the proposal and its ground truth counterpart.}
Building on this idea later works adopt similar feature grouping strategies but explore different network architectures to improve segmentation accuracy and efficiency, such as the recurrent-based network used in RPMNet~\cite{yan2019rpm} in capture temporal relationships and graph convolutional network used in \citet{abdul2022learning} to learn the hierarchical part structure. 
While these methods demonstrate effective articulated part prediction across object categories, they heavily rely on large-scale labeled point cloud datasets for training. 

Another line of work focuses on pose estimation for the articulated parts from a single state of the object, where part segmentation is a necessary first step.
Early works make the assumption that the object category is known and all the instances in the category share the same articulated structure.
Under this assumption, \citet{li2020category} proposed to represent each class of objects in an articulation-aware normalized coordinate space hierarchy (ANCSH) (as shown in \Cref{fig:acnsh_rep}).
The network is trained to map the new object instance to the normalized coordinate space with a branch for assigning the part labels, which are supervised by the per-point ground truth labels.
This method inspired a series of works~\cite{weng2021captra,liu2022toward} that leverage this idea to address the articulated part segmentation and pose estimation on the category level.
The main limitation of these methods is that the assumption of shared part structure across instances in the same category may not hold for all object categories, which limits the generalization capability of the model.
To address this limitation, \citet{geng2023gapartnet} proposed to address the same problem in a category-agnostic manner by introducing generalizable actionable parts (GAPart).
Using these ground-truth GAPart labels as supervision, the trained part segmentation model is more generalizable to unseen object categories.

\input{figures/4_acnsh_rep}

In an effort to reduce reliance on articulation annotation available,
Banana~\cite{deng2023banana} designs a network to be aware of the inter-part equivariance so that the knowledge acquired from objects only in resting states can be generalized to any other articulated states.
This strategy allows the datasets compiled from static objects with segmentation labels to be usable for articulated part perception.
Another line of work in this category attempts to perceive the part structure from RGB(-D) images.
Given a pair of segmented parts in a single-view RGB-D image, \citet{zeng2021visual} proposed FormNet to estimate connectedness and motion flow between the parts in a supervised way.
OPD~\cite{jiang2022opd}, OPDMulti~\cite{sun2023opdmulti} and \citet{wang2025active} propose to learn 2D segmentations from single-view RGB images in a supervised manner, where the ground truth masks are available from the datasets they collected. 
\citet{qian2022understanding} first proposed to infer mobility from an RGB video observing the human-object interaction.
They train a network to regress 3D plane parameters to represent the articulated parts by leveraging the supervision from the 3D datasets without articulation annotation.

For the task of articulated object creation, supervised learning methods are also widely adopted.
A-SDF~\cite{mu2021asdf}, \citet{wei2022self}, and CARTO~\cite{heppert2023carto} propose a line of work for the object-level reconstruction of a shape that is deformable to a specific state.
The model training is supervised by either ground truth SDF or multi-view RGB images.
These supervision signals guide the model to learn an underlying distribution of the geometry that is associated with the articulation states.
To further achieve part-level reconstruction, Ditto~\cite{jiang2022ditto} requires additional supervision from the part segmentation labels on point clouds, and CLA-NeRF~\cite{tseng2022cla} and Real2Code~\cite{mandi2024real2code} learn from segmentation masks on multi-view images.

\mypara{\jy{Semi-weakly supervised learning methods.}}
\jy{While part semantics do not always align with mobility, semantic segmentation labels can serve as a valuable source for refining articulated part segmentation.
Inspired by this idea, \citet{liu2023semi} contributes a semi-weakly supervised approach to tackle the task of object kinematic motion prediction problem.
Their method leverages the fine-grained and hierarchical part labels from PartNet~\cite{mo2019partnet} as weak signals to train a graph neural network. It also prunes the input semantic hierarchy and extracts a mobility tree to represent the object’s articulated structure.
Then parts predicted to have motion in the output mobility tree are used as pseudo-labels to expand the training data in the PartNet-Mobility~\cite{xiang2020sapien} dataset for a part mobility prediction network in a semi-supervised manner.
This semi-weakly supervised pipeline demonstrates its advantage by improving model performance through the use of semantic annotations, which are generally more accessible than motion-part annotations.
However, its effectiveness is limited by the granularity of the semantic hierarchy in the source dataset, as a coarse hierarchy may lead to imprecise mobility predictions.
Additionally, the method relies on a sufficient amount of ground-truth labeled data to train the model before generating pseudo-labels for each category, as illustrated in \Cref{fig:semi_weak}.
The reliability of these pseudo-labels depends on the quality of the motion prediction network, and errors in motion estimation can introduce noise into the training process, potentially affecting overall model performance.}

\input{figures/4_semi_weakly}

\mypara{Self-supervised learning methods.}
Since the articulated part segmentation and articulation annotations are expensive to obtain, self-supervised learning methods have been proposed recently to reduce the reliance on these supervision signals.
Taking a point cloud with a known number of parts as input, \citet{liu2023self} proposes to learn the segmentation by extracting part-level equivariant features and factorizing the shape into canonical states.
Through the iterative processing of canonicalization and then reconstruction using the factorized parameters, part decomposition can be learned by minimizing the reconstruction error on the input point cloud.
PARIS~\cite{liu2023paris}, \citet{weng2024neural}, and ArtGS~\cite{liu2025building} reconstruct articulated objects with part structure from a pair of multi-view RGB(-D) images that observe the object in two different states by leveraging differential rendering.
By identifying the static and moving components through iterative optimization, the part-level geometry can be reconstructed by minimizing the photometric error between the input images and the rendered ones from the reconstructed surface.
Following the trend, LEIA~\cite{swaminathan2024leia} proposes to reconstruct the interpolated object between different input states by leveraging a hypernetwork to modulate the network parameters of the reconstructed NeRF.
They encode each state of the object into a latent code which can be linearly interpolated to generate the latent for the intermediate state, which is taken as the input to the hypernetwork to modulate the NeRF.
This optimization process is supervised by the photometric loss to ensure the consistency of the reconstructed object and several regularization terms to enforce a smooth and continuous latent manifold. 
REACTO~\cite{song2024reacto} and RSRD~\cite{kerr2024robot} follows a similar idea to reconstruct the articulated objects from videos observing the object in motion.
A photometric loss is used in these methods to learn the neural fields that represent the object with part-level structure.

\input{figures/4_ddpm_nap.tex}

\mypara{Generative models.}
Leveraging the recent advance of diffusion models and autoregressive models in 3D generation, a line of work has focused on training generative models to synthesize articulated objects with compositional part structures.
NAP~\cite{lei2023nap} pioneers a novel representation that enables diffusion models to generate articulated objects with variational structures.
As the diffusion model typically requires a fixed-size input, NAP represents the object as a set of parts padded to a fixed number, where each part is described by shape-motion attributes and its geometry encoded as a latent shape code, as shown in \Cref{fig:ddpm_nap}.
During generation, all parts are fully connected, and the model jointly synthesizes part attributes while reasoning about their dependencies to form a kinematic tree.
The training follows a denoising diffusion process, where the model minimizes residual noise loss at each step to iteratively refine the generated structure.
Once trained, new articulated object samples can be generated by progressive denoising from random noise.
The denoiser is built on a graph convolutional network (GCN) architecture, which helps model hierarchical part relationships and capture kinematically plausible articulation structures.
\jy{This framework showcases the effectiveness of diffusion models for articulated object generation, but the model shows limitations in user controllability over the generation process.}

CAGE~\cite{liu2024cage} \jy{and SINGAPO~\cite{liu2024singapo}} share a similar spirit to train diffusion models but focus on improving the controllability on the structure level and the user interaction in the generation process.
These models tokenize part attributes and adopt a transformer-based architecture in the denoising network to capture part relationships via attention mechanisms.
A key innovation is decoupling the part dependency graph from the output and using it as a guidance mechanism to enforce structural consistency throughout the generation process.
CAGE proposes to use varying masking strategies on self-attention layers to enforce the graph conditional input effectively.
\jy{Building on top of CAGE, SINGAPO introduces additional cross-attention blocks to inject spatial information into the part structure from an image feature map.
This enables the model to generate multiple plausible articulated objects that are all consistent with an input image, improving the flexibility of image-conditioned generation. 
These methods demonstrate the potential of diffusion models for generating articulated objects with controllable structures and user-driven interactions. 
However, due to their part retrieval-based formulation, they lack the ability to synthesize diverse part geometries or ensure high compatibility between generated parts and the input image.
Future work could explore integrating shape generation under a structural context, enabling more expressive and detailed synthesis to better align with input constraints and user intent.}

\input{figures/4_meshart}

\jy{Along with the generative approach built on diffusion models, another line of work explores generating articulated parts in an autoregressive manner.
MeshArt~\cite{gao2024meshart} explores the articulated mesh generation in a part-by-part fashion.
The process begins by generating a high-level articulation-aware object structure, which then guides the generation of individual part meshes.
The key insight is to represent both structural information and part geometry as sequences of quantized triangle embeddings, making them well-suited for a unified autoregressive framework that captures articulation and fine-grained geometry jointly.
This design allows the model to generate highly detailed meshes with realistic articulation, but it comes with certain limitations.
The model is trained per object category, restricting its generalization across diverse articulated objects.
Additionally, it assumes that articulation dependencies are limited to direct parent-child relationships (i.e., within one hop in the kinematic tree), which may not fully capture the complex hierarchical articulation structures present in more intricate objects.
Similarly, ArtFormer~\cite{su2024artformer} adopts an autoregressive strategy to iteratively generate child parts for each known part.
Each part is represented as a separate token with high-dimensional features, and feature chunks can be decoded into shape and articulation parameters.
On the shape side, each part is represented by a bounding box and a latent code which is trained to be decoded into SDFs.
A key advantage of this tokenization strategy is its simple alignment with text and image tokens, enabling the model to generate diverse articulated objects from user-provided textual or visual descriptions.
This capability bridges text-to-3D synthesis with articulated object generation, offering a more interactive and controllable approach to structured 3D content creation.}

\mypara{\jy{Transfer learning methods.}}
\jy{Trained on vast amounts of data, foundation models such as large language models (LLMs) and vision-language models (VLMs), along with general-purpose vision models like SAM~\cite{kirillov2023sam} and DUSt3R~\cite{wang2024dust3r}, have demonstrated significant potential in enhancing downstream tasks. 
These models serve as a base that can be adapted to a wide range of tasks with fine-tuning or prompting with a few examples.
A recent line of work leverages these models to advance articulated object creation, utilizing their capabilities in understanding, segmenting, and reasoning about complex structures.
\citet{liu2023few} tackles the problem of few-shot articulated mesh generation by fine-tuning a convex deformation network pre-trained on part convex decomposition of the objects from similar categories.
Real2Code~\cite{mandi2024real2code} proposes a framework to reconstruct articulated parts from multi-view images by fine-tuning SAM for segmenting articulated parts, and then lifting each part to a 3D point cloud using DUSt3R.
Articulate-Anything~\cite{le2024articulate} proposes an actor-critic system to generate articulated part configurations using programs.
Multiple VLM and LLM agents are prompted to specialize in different subtasks, such as object detection and reasoning from the visual input, part mesh retrieval and arrangement, rating the realism of the output, and providing specific feedback to correct the output.
DragAPart~\cite{li2024dragapart} generates images of articulated objects constrained by a drag action by fine-tuning the stable diffusion model~\cite{rombach2022high} with paired drag-annotated images.
Although trained on synthetic images, the model can generalize well to real images by leveraging the text-to-image prior in the pre-trained diffusion model.
Articulate AnyMesh~\cite{qiu2025anymesh} proposes a pipeline to segment the input mesh by aggregating 2D segmentation results from multi-view images.
The 2D articulated part segmentation is achieved by few-shoting the PartSlIP++ model~\cite{zhou2023partslip} using a few example masks for each category.
It effectively leverages the advanced image segmentation models to assist the 3D segmentation task on surface meshes.
For the occluded geometries, a refinement step is followed, which distills 3D geometry-texture priors from a pre-trained RichDreamer model~\cite{qiu2024richdreamer} using SDS loss~\cite{poole2022dreamfusion}.}

\jy{These transfer learning based methods typically design a multi-stage pipeline that brings together the strengths of different foundation models to tackle each subproblem in the object creation process.
Benefiting from powerful prior knowledge encoded in the base modules, these methods can potentially reduce the need for extensive annotated data to train a model from scratch, but only require a few examples to prompt the model or even used in zero-shot settings.
However, the performance of these methods is highly dependent on the generalization capability and reliability of the base models.
How robust the model can be adapted to the target task and how well it can generalize to unseen data are still open questions that need further exploration.}

%% file: figures/4_mobtree_sharf.tex
\begin{figure}[t]
    \begin{center}
    \includegraphics[width=\linewidth]{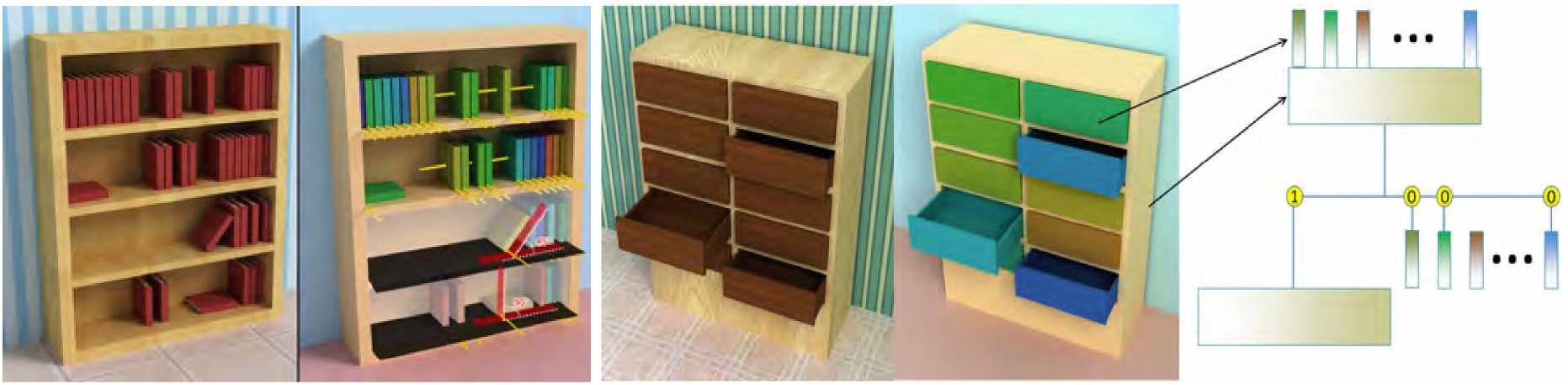}
    \end{center}
    \caption{ Method proposed by \citet{sharf2014mobility} for object/part segmentation from a mesh input by relying on the objects/parts being observed in multiple states or poses.
    Figure reproduced from original paper~\cite{sharf2014mobility}.}
    \label{fig:mobtree_sharf}
\end{figure}

%% file: figures/4_arti_gen.tex
\begin{figure*}[t]
    \begin{center}
    \includegraphics[width=\linewidth]{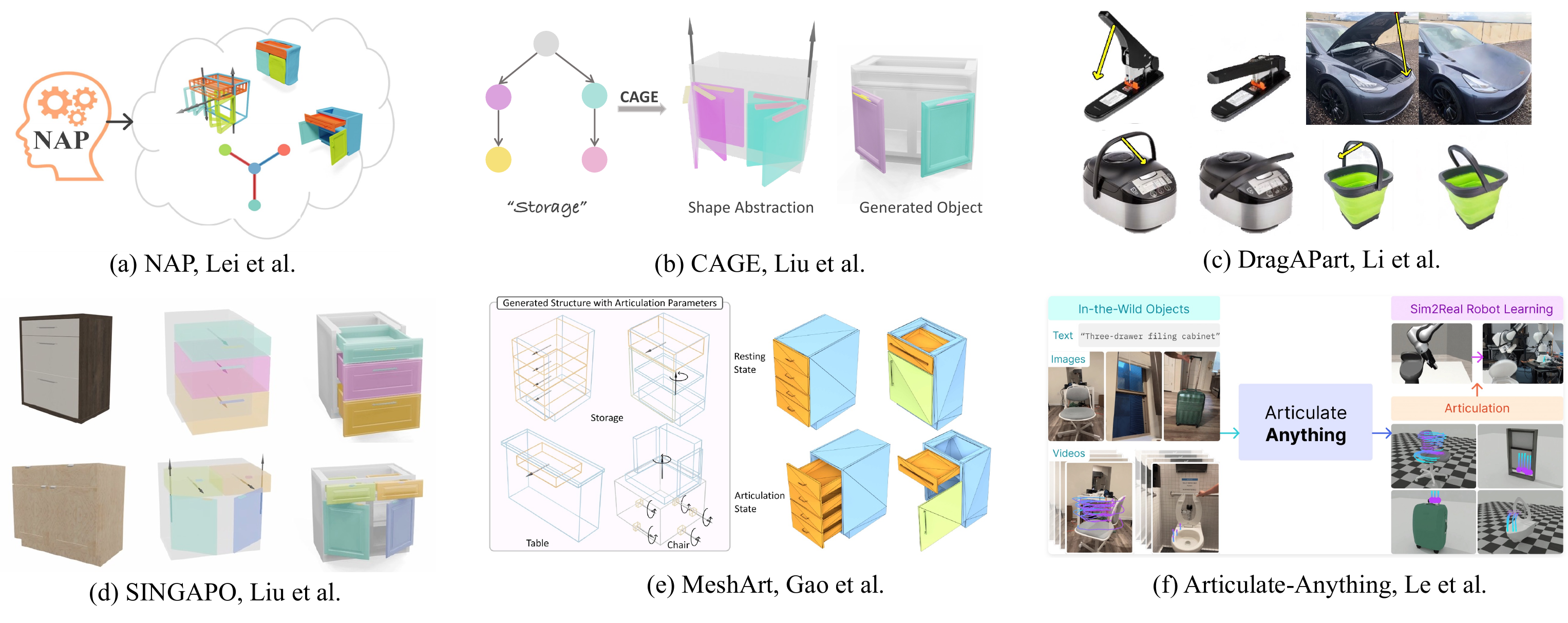}
    \end{center}
    \caption{Representative methods for articulated object generation.
    Figure reproduced from original papers~\cite{lei2023nap,liu2024cage,li2024dragapart,gao2024meshart,liu2024singapo,le2024articulate}.}
    \label{fig:arti_gen}
\end{figure*}

%% file: figures/4_interm_rep.tex
\begin{figure}[t]
    \begin{center}
    \includegraphics[width=\linewidth]{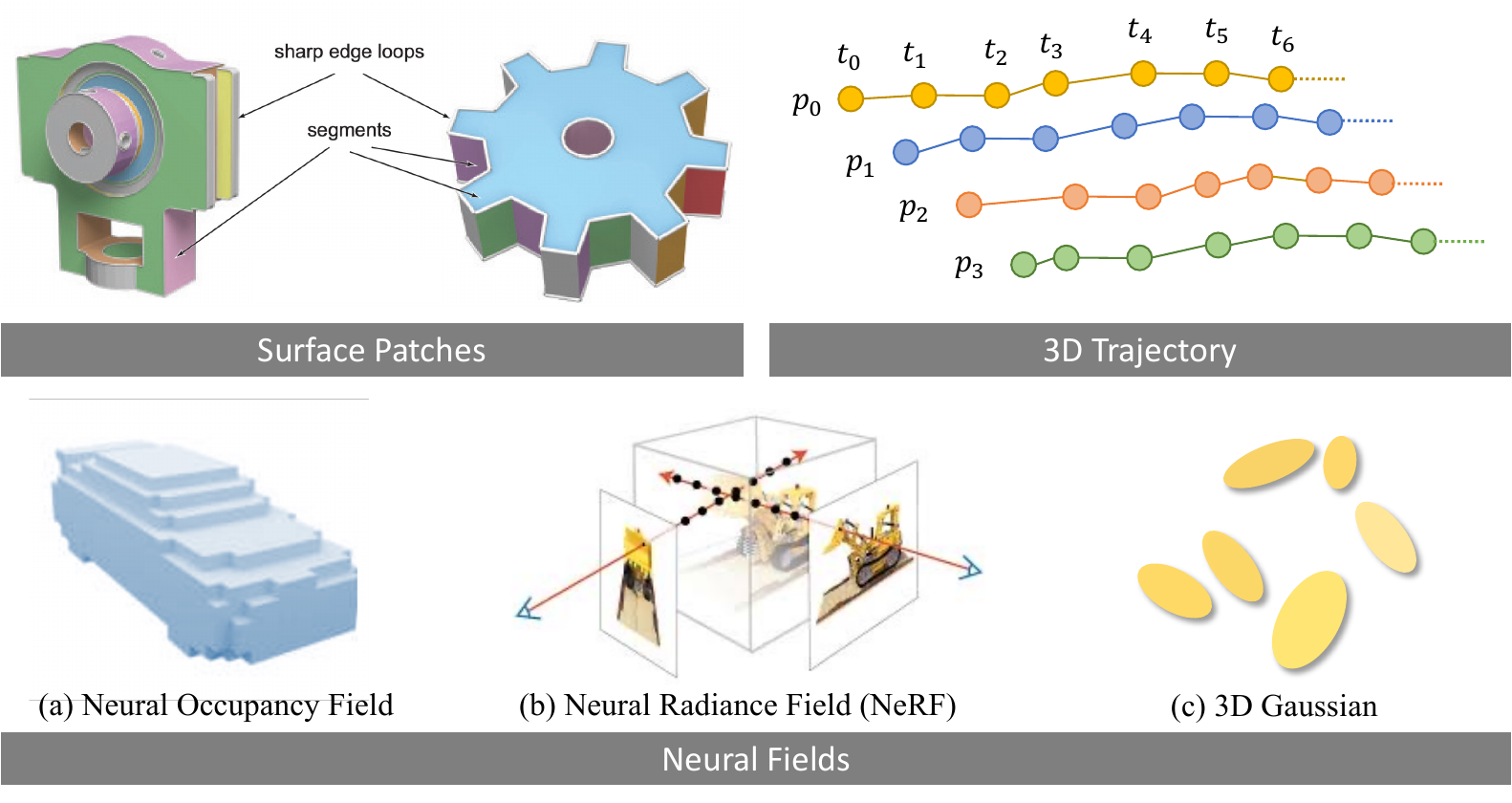}
    \end{center}
    \caption{ Selected examples of the intermediate representation used for articulated part perception and creation.
    Figures reproduced from the original papers~\cite{mitra2010illustrating,li2016mobility,mescheder2019occnet,mildenhall2021nerf,kerbl20233d}. 
    }
    \label{fig:interm_rep}
\end{figure}

%% file: figures/4_acnsh_rep.tex
\begin{figure}[t]
    \begin{center}
    \includegraphics[width=\linewidth]{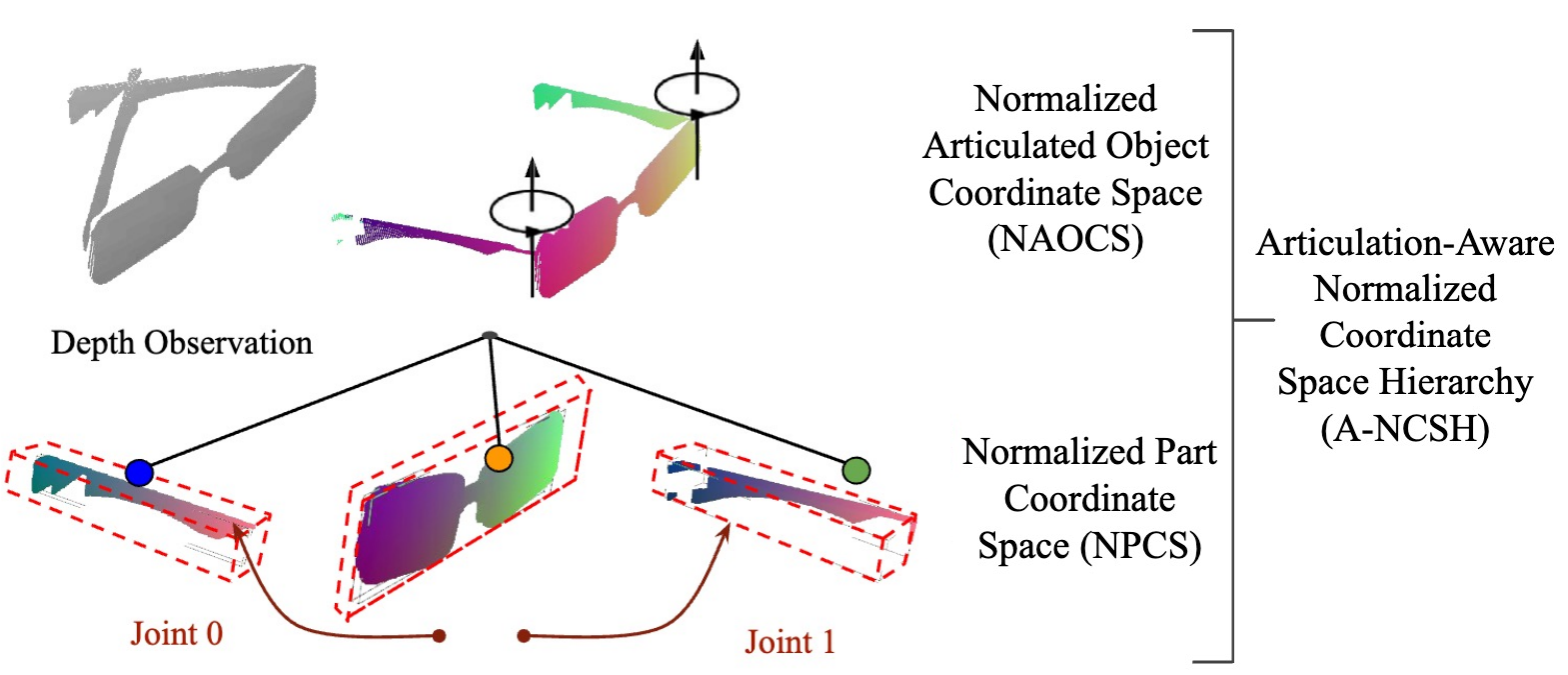}
    \end{center}
    \caption{Articulation-aware Normalized Coordinate Space Hierarchy (ANCSH) proposed by \citet{li2020category}.
    ANCSH is a category-level representation that defines canonicalized object state with joint parameters in NAOCS, and normalizes each part in NPCS while maintaining the part orientation in NAOCS.
    The colors represent the corresponding coordinates in each space.
    Figure reproduced from original paper~\cite{li2020category}.}
    \label{fig:acnsh_rep}
\end{figure}

%% file: figures/4_semi_weakly.tex
\begin{figure}[t]
    \begin{center}
    \includegraphics[width=\linewidth]{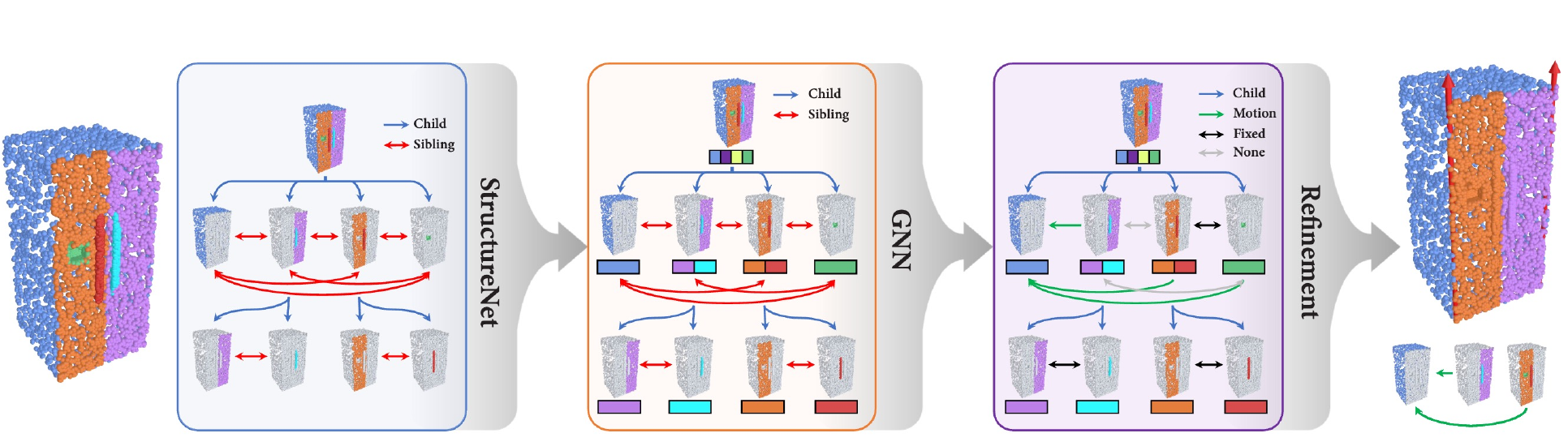}
    \end{center}
    \caption{\citet{liu2023semi} introduces a semi-weakly supervised learning approach that leverages the fine-grained and hierarchical part labels as weak signals to predict mobility trees, which are then used as pseudo-labels to train mobility analysis models.
    Figure reproduced from original paper~\cite{liu2023semi}.}
    \label{fig:semi_weak}
\end{figure}

%% file: figures/4_ddpm_nap.tex
\begin{figure}[t]
    \begin{center}
    \includegraphics[width=\linewidth]{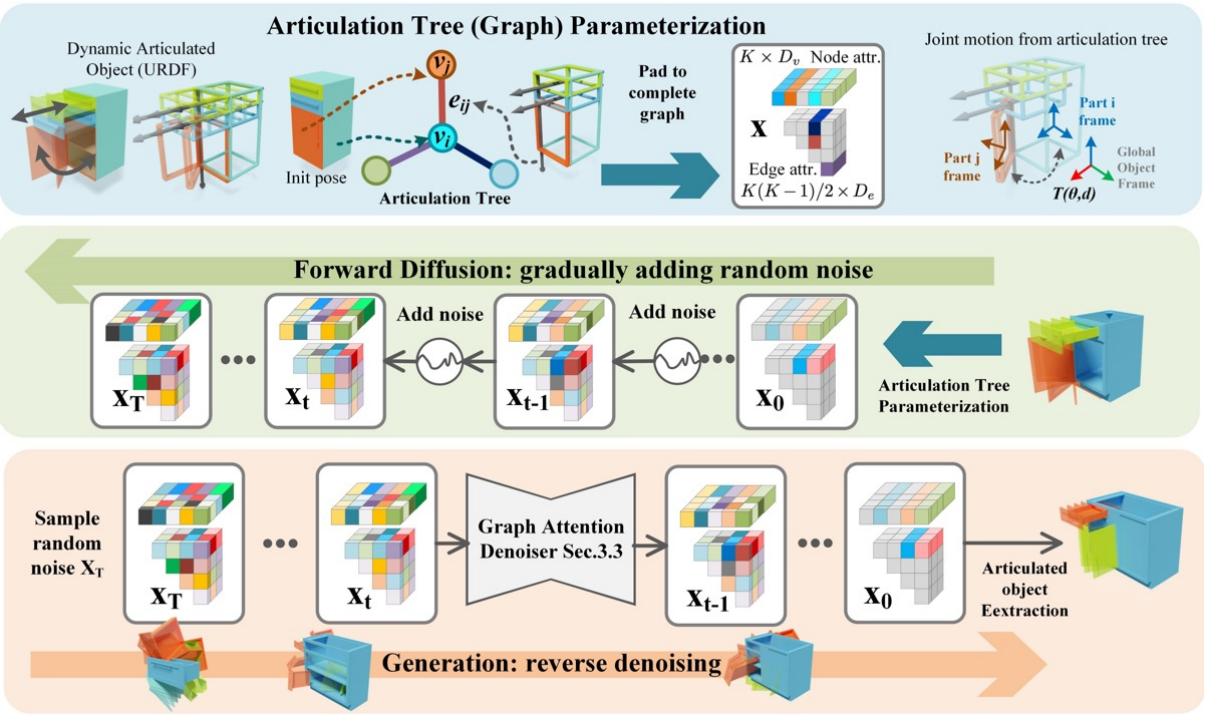}
    \end{center}
    \caption{Generative architecture based on diffusion model for unconditional articulated object synthesis from NAP~\cite{lei2023nap}.
    Figures reproduced from the original paper~\cite{lei2023nap}.}
    \label{fig:ddpm_nap}
\end{figure}

%% file: figures/4_meshart.tex
\begin{figure}[t]
    \begin{center}
    \includegraphics[width=0.8\linewidth]{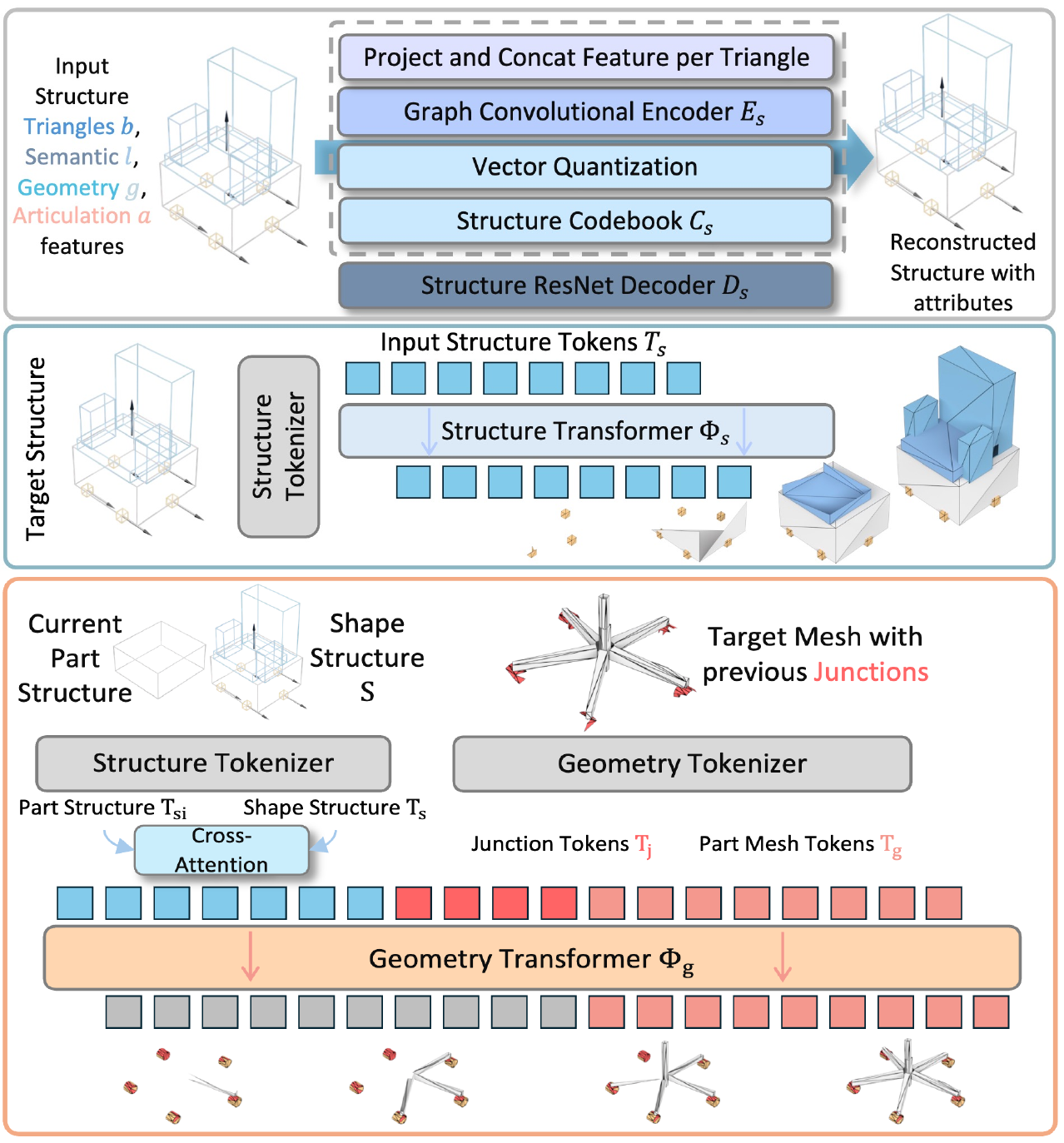}
    \end{center}
    \caption{Autoregressive model for mesh-based articulated object generation from MeshArt~\cite{gao2024meshart}.
    Figures reproduced from the original paper~\cite{gao2024meshart}.}
    \label{fig:meshart}
\end{figure}

%% file: secs/5_articulation.tex
\section{Articulation Modeling}
\label{sec:articulation}
The articulation model of an object is a representation for describing the part mobility and the relationship between the parts.
Understanding the kinematic structure of an object is a shared goal for both tasks of articulated part perception and articulated object creation.
One of the key challenges in articulation modeling is how to deal with objects with different kinematic structures, such as the number of joints, the type of joints, the degrees of freedom (DoF) of the joints, and what hierarchy the joints are organized.
In~\Cref{tab:arti}, we summarize the related works from the perspective of articulation modeling in terms of the assumptions made on the articulation model, the representation of articulated motion, and the methodology used for articulation modeling.
In the following, we will discuss each aspect in detail.

\input{tabs/articulation}

\subsection{Representations of Articulated Motion}

\mypara{Joint parameters.}
The most common way to represent part articulation is through joint parameters, which define the motion properties of articulated parts in terms of joint type, axis, state, and limits.
This representation is widely used in robotics and simulation, as it provides a structured way to model how parts move relative to each other.
The joint types most commonly seen are the 1 DoF joints, such as the revolute joint and the prismatic joint.
The joint with more than 1 DoF, such as the ball joint, can also be seen in more complex objects, e.g. mechanical parts.
An overview of the joint types that are covered in the literature is shown in~\Cref{fig:joint_types}.
The joint axis consists of a direction vector for orientation and a position vector for the pivot.
The joint state captures its current configuration, typically as a rotation angle (revolute) or translation distance (prismatic).
Joint limits define the allowed motion range to ensure physical plausibility.

The line of work on mobility analysis formulates joint parameter estimation as a combination of classification and regression problems, depending on the task requirements.
Typically, joint type, position, and orientation are the primary components to be estimated. 
Some works additionally estimate the joint limit~\cite{xu2009joint,mitra2010illustrating,hu2017learning,li2020category,shi2021self,qiu2025anymesh}, which can be more challenging as it requires the network to understand the physical constraints of the object.
Other works focus on estimating the pose of the articulated parts from 3D observation of objects~\cite{abbatematteo2019learning,li2020category,weng2021captra, liu2022toward, liu2023self,geng2023gapartnet}. 
Part pose is typically represented as a 3D amodal bounding box with a free 6D pose. 
By mapping the estimated pose to a normalized coordinate space introduced by \citet{li2020category}, each part pose can be further converted into the joint state.

\mypara{Kinematic tree.}
The kinematic tree serves as a hierarchical representation of the joints and parts of an articulated object. 
This structure is particularly crucial for depicting objects with multiple parts and joints, as it clearly outlines the dependencies between them
An example of the kinematic tree is shown in~\Cref{fig:arti_rep}.
Extracting the kinematic tree is a challenging problem since the kinematic structure varies from object to object with different numbers of parts and is structured in different topologies.
Some existing works bypass this step by assuming the kinematic tree is predefined per object class~\cite{abbatematteo2019learning,li2020category,liu2023self,mu2021asdf,wei2022self}, or by estimating only one joint at a time~\cite{hu2017learning,jain2021screwnet,jiang2022ditto,tseng2022cla,heppert2023carto,liu2023paris}.
And a large portion of the works dealing with multiple joints also do not represent the kinematic tree in their model~\cite{yi2018deep,wang2019shape2motion,yan2019rpm,shi2021self,sun2023opdmulti,geng2023gapartnet} as they assume that all the articulated parts move relative to the static part.

There are only a few works that explicitly model the kinematic tree as the output of the mobility analysis.
The early works by \citet{mitra2010illustrating} and \citet{sharf2014mobility} extract the kinematic tree from the input mesh.
They share a similar idea to leverage the intersection of the part surfaces to identify the connection between the parts.
\citet{mitra2010illustrating} detected the contact area between the parts to determine the topology of the interaction graph, and \citet{sharf2014mobility} applied a segment-cluster process to compute the support-tree hierarchy from decomposed parts.
Later works by \citet{abdul2022learning} and \citet{liu2022toward} attempt to predict the kinematic tree from the input point cloud by labeling the dependency existence between each pair of segmented parts. 
\jy{Articulate AnyMesh~\cite{qiu2025anymesh} predicts the kinematic dependency using VLMs.}
For the work focusing on articulated object generation~\cite{lei2023nap,liu2024cage,liu2024singapo,le2024articulate,gao2024meshart,su2024artformer}, the kinematic tree is an essential output, as their ultimate goal is to synthesize fully articulatable assets that can be directly integrated into the simulation environment.

\mypara{Deformation flow.} 
Deformation flow is a general form of motion representation that can describe not only the rigid motion but also the motion as free-form deformation, making it highly versatile for modeling articulation. 
It describes the motion as a vector field, where each vector denotes the spatial displacement for each point in space.
Unlike traditional joint-based representations, deformation flow allows for more flexible motion modeling, especially when the motion involves non-trivial transformations.

For mobility analysis, \citet{yi2018deep} and RPM-Net~\cite{yan2019rpm} opt to use deformation flow to represent the articulation motion.
They both leverage recurrent neural networks, such as LSTM, to directly predict the deformation flow from the input point cloud.
An example of the deformation field prediction network is illustrated in~\Cref{fig:rpmnet}.
One of the advantages of using deformation flow is that it can handle the motion of the parts in a free-form manner.
This flexibility is useful when the number of parts is unknown as the deformation on each point can be later grouped to form articulated parts in arbitrary numbers.
The deformation flow can be particularly useful when modeling the non-trivial motion.
For example, RPM-Net can model the opening and closing motion of the cover of an umbrella, which is a non-linear motion that cannot be represented by a simple rigid transformation.

Similarly, A-SDF~\cite{mu2021asdf}, \citet{wei2022self}, and REACTO~\cite{song2024reacto} also leverage the deformation field to represent articulation for the objects in the task of articulated object creation.
These methods optimize coordinate-based networks to implicitly represent the deformation, conditioned on the articulation state of the object.
Once trained, the network allows the reconstructed object to be deformed into different states by querying the neural field at different timesteps, enabling continuous articulation modeling.

However, a key limitation of deformation flow or fields is that they are less constrained than strict rigid transformations, which can be problematic when dealing with rigidly moving parts.
To alleviate this issue, \citet{yi2018deep} proposes to optimize the deformation flow with an as-rigid-as-possible (ARAP) objective~\cite{sorkine2007rigid} to preserve the local rigidity of the parts.
\jy{RSRD~\cite{kerr2024robot} estimates articulation by tracking the pose for each cluster of 3D Gaussians over time in an input video.
By treating 3D Gaussians as soft points in space, an ARAP loss penalizes deviation between neighboring Gaussians, ensuring that deformation follows locally rigid motion.
This approach bridges deformation-based articulation modeling with Gaussian-based representation learning, demonstrating the potential for more flexible yet structured articulation tracking.}

\jy{\mypara{Discussion on articulation representations.}}
\jy{Joint parameters, kinematic trees, and deformation fields are three common ways to represent articulation, each with its own strengths and challenges. 
Joint parameters are fundamental for describing discrete motion constraints and are widely used in robotics and simulation.
Kinematic trees define the hierarchical structure of articulation dependency, which complements joint parameters to form a complete articulation model while also serving as a crucial component for simulation-ready articulated assets.
While often overlooked due to the simplicity of many articulated objects with single-level structures, kinematic trees become essential when handling objects with deeper articulation hierarchy (e.g., table lamps with multiple arms, robotic mechanisms, mechanical assemblies). 
Modeling such structures introduces additional challenges for mobility analysis and articulated asset creation, as it requires reasoning over highly structured data.
In contrast, deformation fields offer a continuous and flexible representation, capable of capturing both rigid and non-rigid motion.
However, they lack explicit structural constraints when modeling piece-wise rigidity, usually requiring additional regularization techniques to enforce local rigidity.
}

\input{figures/5_rpmnet.tex}

\subsection{Assumption on the Articulation Model}
There are several axes of assumptions made on the articulation model by the existing works.
These assumptions usually are made to simplify the problem to make it more tractable.
We summarize these common assumptions to show the research progress below.

\mypara{The number of degrees of freedom (\# DoF)}.
Most of the existing works assume to only deal with the 1 DoF joints, including the revolute joint, the prismatic joint, and the helical joint.
These joints are the most common types of joints that are seen in the daily articulated objects, such as cabinets, dishwashers, refrigerators, etc.
And their motion is constrained to be linear and easily represented by a combination of rotation and translation constrained by a single joint axis.
When dealing with more complex objects, such as delicate mechanical systems, the joint with more than 1 DoF, such as the ball joint, can also be seen.
A few methods~\cite{xu2009joint,mitra2010illustrating,li2016mobility,yi2018deep} have been proposed to deal with joints up to 3 DoF, which is achieved by either leveraging the deformation field to represent the motion or by combining with the handcrafted features on the part geometry to classify the joint type.

\mypara{Known object category (Obj. cat.)} 
As mentioned above, one of the main challenges in articulation modeling is the diversity of the kinematic structure of the objects.
One way to generalize the knowledge learned from the training data to the unseen objects is to build a category-level model, which is based on the assumption that the objects in the same category share the same structures.
The structure implied by this line of work~\cite{abbatematteo2019learning,li2020category,liu2023self} is not only about the kinematic structure but also the geometric structure.
It means that the number of parts, how the parts are arranged spatially, and the way each part is articulated are all shared among the objects in the same category.
This presents a strong prior for the model to learn from.
In some cases, this level of category can be aligned with the semantic class of the objects whose kinematic structure is relatively simple and fixed, such as eyeglasses, laptops, scissors, etc.
However, this assumption is highly constrained since this is not the case for more objects with more complex structures.
For example, a refrigerator with two doors stacked vertically and a refrigerator with two doors stacked horizontally are considered to be from the same semantic class but with different structures.
A separate model should be trained for each kind of refrigerator under this assumption.
It makes the model less scalable to real-world applications with a diversity of objects.

\mypara{Known number of joints (\# Joints)}.
Whether the number of joints is known is another assumption made by the existing works.
The previous assumption of knowing the object category inherently implies that the number of joints is known.
There are also a few works~\cite{hu2017learning,jain2021screwnet,jiang2022ditto,tseng2022cla,heppert2023carto,liu2023paris} that assume to estimate only one joint at a time but not knowing the motion the joint exhibits.
It can be useful to generalize the method across objects with varied kinematic ways for simple objects.
However, it is not scalable to more complex objects with more than one joint, highlighting a limitation in this line of work.

\subsection{Methodologies for Articulation Modeling}
\label{subsec:arti_strategy}
\mypara{Handcrafted methods}.
Early works leverage non-learning-based methods to infer the articulation model.
\citet{xu2009joint} applied slippage analysis to compute a set of joint parameters from a mesh input.
The output includes the number of rotational and translational degrees of freedom which is thresholded to be up to three, and the corresponding axes for each slippable motion.
To determine the motion range for each joint, they design a trial-and-error bisection process to decide the feasibility.
They estimate the joint limit by probing the possible motions until the motion results in an unrealistic geometry, e.g. penetration occurs between the parts.
\citet{mitra2010illustrating} define a list of heuristic rules to determine the joint type, axis, and limit for each mechanical part.
Once detecting the supporting-supported relationship between parts, \citet{sharf2014mobility} extract this connectivity to form a kinematic tree.
By leveraging the reoccurrence of the objects and parts in the scene with different poses, they design rules to determine the motion type and axes by using the PCA axes of the part geometry.
Taking a point cloud sequence as input, \citet{li2016mobility} propose a random sampling consensus method in 4D to fit motion on 3D trajectories.
Then these motions can be clustered to convert to the joint parameters mathematically and the joints can be organized in a mobility graph.
CLA-NeRF~\cite{tseng2022cla} assumes the object has a single revolute joint whose axis is at the intersection of the two articulated segments.
So once the parts are segmented in the NeRF, the joint axis can be directly voted by the points near the intersected area by applying linear regression.

\input{figures/5_acnsh.tex}

\mypara{Supervised learning}.
A recent line of work focusing on mobility analysis leverages supervised learning to predict the articulation model from a single state of the object.
Depending on the articulation representation chosen and perception input, the training process can be supervised by the ground truth motion parameters~\cite{wang2019shape2motion,jain2021screwnet,qian2022understanding,nie2023sfa} or point displacements~\cite{yi2018deep,yan2019rpm}.
\jy{As a representative work shown in \Cref{fig:shape2motion}, Shape2Motion~\cite{wang2019shape2motion} designs a motion attribute proposal module to decode point-wise features into parameters for all the possible articulation in the object.
Rather than directly regressing the joint axis, they reformulate joint location prediction as a point selection problem and the joint orientation problem as a classification task over 14 discrete directions, which reduces the complexity of direct regression in high-dimensional continuous spaces.
Once each motion is proposed, a matching module follows to pair articulations with the corresponding parts, allowing for cases where a single articulated part may exhibit multiple 1-DoF motions.}
Different from the above works that explicitly use motion annotation, \citet{hu2017learning} propose to classify the motion type from a mesh input through a metric learning model.
The training is supervised by a distance function with several constraints to pull the object pairs with the same motion type closer and push the ones with different motion types away.

\jy{Another line of work tackles the task of articulated part pose estimation from the depth image of an object in an arbitrary articulation state in a supervised manner.
By assuming the articulation structure is shared within each object category, \citet{li2020category} pioneers in designing an articulation-aware normalized coordinate space hierarchy (ANCSH), which is used to canonicalize the perceived object at the part level.
As shown in \Cref{fig:acnsh_rep}, ANCSH is composed of a Normalized Articulated Object Coordinate Space (NAOCS) on top of a set of Normalized Part Coordinate Spaces (NPCSs) per part, and the joint parameters in the resting state have been pre-defined in NAOCS.
Once this hierarchical representation is defined for each object category, a deep neural network (illustrated in \Cref{fig:acnsh}) is trained to predict the mapping from each point to a coordinate in NPCS, followed by a head to predict the transformation from each NPCS to NAOCS and another head to predict the point-joint association and joint parameters in NAOCS.
The training is supervised by multiple losses using ground truth outputs, including per-point NPCS and NAOCS mapping, per-part transformations, joint parameters, etc. 
Leveraging this ANCSH representation, CAPTRA~\cite{weng2021captra} and GAPartNet~\cite{geng2023gapartnet} extend the problem setting to pose tracking and domain-generalizable object perception.
}

\mypara{Self-supervised learning}.
Although supervised learning is effective in learning the articulation model, a large amount of labeled data is required to train the model.
To alleviate this issue, a few works~\cite{shi2021self,huang2021multibodysync,liu2023self,lin2024autourdf} propose to leverage self-supervised learning to estimate the articulation model by leveraging the correspondence across multi-state observations of the object.
\citet{shi2021self} leverage a network to learn the point correspondence across different frames of the point cloud sequence from the input.
Once the 3D trajectories are constructed by learning the consistent point correspondence, the joint parameters can be estimated by fitting the motion on the trajectories.
One limitation of this work is that it requires observing a motion sequence as input, which usually involves human intervention while capturing the data.
This tedious motion-capturing process makes the method less scalable to real-world applications.
To alleviate this issue, \citet{liu2023self} propose to learn the joint parameters from a single point cloud in a self-supervised manner.
By factorizing each articulated part to its own normalized canonical space and then reconstructing it back to the input state using an optimized transformation, the reconstruction consistency in this process can be used to supervise the training.
However, this work is limited by the assumption that the model is category-specific, meaning the geometric and kinematic structure of the object is known in advance.

\input{figures/5_paris}

In the task of articulated object reconstruction, PARIS~\cite{liu2023paris} proposes to learn the joint parameters from a pair of multi-view RGB images in a self-supervised manner. \jy{As illustrated in \Cref{fig:paris}, the estimation of the joint parameters is supervised by the consistency of a photometric loss across two articulated states given as input.
Although this method can be generalizable across object categories and does not rely on any 3D or articulation supervision, it assumes only one part is moving in the input observation.}
\citet{weng2024neural} and ArtGS~\cite{liu2025building} follows a similar idea to optimize the joint parameters from a pair of observations in multi-view RGB-D images.
Reconstruction losses on the RGB-D frames are used to first segment the parts in two states of the object, and then the joint parameters can be computed by optimizing the point correspondences between the segments from the two states.
RSRD~\cite{kerr2024robot} models the object dynamics from human demonstration videos using a photometric consistency loss and regularization terms to constrain the motion of the parts with as-rigid-as-possible (ARAP)~\cite{sorkine2007rigid} penalty.
This work tracks the articulated part in 4D poses along the time steps, so it can be transferred to a physical grasping robot to execute these motions.

\mypara{Generative methods}.
As mentioned in \Cref{subsec:geo_strategy}, there is a line of work~\cite{lei2023nap,liu2024cage,liu2024singapo,gao2024meshart,su2024artformer} that trains generative models from scratch to learn the joint distribution of geometry and articulation for each part.
On the articulation side, all existing methods model the motion properties explicitly through joint parameters.
The most common approach~\cite{liu2024cage,liu2024singapo,gao2024meshart,su2024artformer} is to parameterize joints with joint location, orientation, and motion limits, with some methods also including joint type~\cite{liu2024cage,liu2024singapo} and a joint existence indicator~\cite{lei2023nap,gao2024meshart}.
NAP~\cite{lei2023nap} adopts Plucker coordinates for joint representation, using a unit direction vector $l$ and momentum $m$ perpendicular to $l$, providing a unified representation for prismatic and revolute joints.
Notably, all the above joint parameters are defined in the global object frame, which is consistent with the part shape representation.
During training, the loss function is designed to enforce consistency with the ground-truth joint parameters per part.

\jy{However, an open question remains as to whether a global-frame or local-frame parameterization is more efficient in the generative context.
The global-frame representation captures the distribution across different objects in a coordinated manner, but it is sensitive to the object's orientation and articulation states.
A more local representation can potentially capture the connection within the part geometry in a more compact way.
Exploring alternative formulations to compactly encode joint parameters alongside part geometry is a promising direction for future research, leading to more scalable and generalizable generative models for articulated objects.}

\input{figures/5_articulate_anything}

\mypara{\jy{Transfer learning methods.}}
\jy{Real2Code~\cite{mandi2024real2code} and Articulate-Anything~\cite{le2024articulate} are two recent works that leverage pre-trained LLMs to transfer language reasoning and code generation capabilities to the articulation understanding problem.
These methods bridge the gap between articulation modeling and generative programming, enabling articulated object representations that are both interpretable and simulation-ready.
Real2Code abstracts the part layout as oriented bounding boxes after segmenting articulated parts. 
It then defines joint parameters relative to these bounding boxes, formulating joint prediction as a classification problem.
Taking the bounding boxes as input, they fine-tune CodeLlama~\cite{roziere2023codellama} to predict the joint parameters and generate code to construct the objects into a compact representation that can be directly used in physics simulations.
In contrast, Articulate-Anything builds a joint prediction system on VLMs.
It takes code representations that configure link placements as input and generates kinematic parameters between parts. 
The prediction is iteratively refined by taking feedback from another program that criticizes the realism of the results by comparing it with the input image or video.
Both approaches demonstrate the potential of LLMs and VLMs in articulation reasoning and code-based object generation.
Articulate AnyMesh~\cite{qiu2025anymesh} also leverages VLMs to analyze the part articulation.
Based on some heuristic rules, they prompt GPT-4o to make selections from several candidate markers on the rendered images or possible joint directions, depending on the joint type.
}

%% file: tabs/articulation.tex
\begin{table*}
    \centering
    \resizebox{\textwidth}{!}{
    \begin{tabular}{@{}lcccccccccll@{}}
    \toprule
    & \multicolumn{3}{c}{Assumption} & \multicolumn{6}{c}{Articulated Motion Representation} & \multicolumn{2}{c}{Methodology}  \\ 
    \cmidrule(l){2-4}\cmidrule(l){5-10}\cmidrule(l){11-12}    
                                                & \# joints & \# DoF        & obj. cat.  & joint type   & joint axis    & joint state   & motion range  & kine. tree    & deform. flow  & strategy          & supervision              \\ \midrule
    \multicolumn{12}{l}{Articulated Part Perception} \\ \midrule 
    \citet{xu2009joint}                         & \xmark    & 3             & \xmark     & \cmark       & \cmark        & \xmark        & \cmark        & \xmark        & \xmark        & handcrafted       & -                        \\
    \citet{mitra2010illustrating}               & \cmark    & 3             & \xmark     & \cmark       & \cmark        & \xmark        & \cmark        & \cmark        & \xmark        & handcrafted       & -                        \\
    \citet{sharf2014mobility}                   & \xmark    & 1             & \xmark     & \cmark       & \cmark        & \xmark        & \xmark        & \cmark        & \xmark        & handcrafted       & -                        \\
    \citet{li2016mobility}                      & \xmark    & 3             & \xmark     & \cmark       & \cmark        & \xmark        & \xmark        & \cmark        & \xmark        & handcrafted       & -                        \\   
    \citet{hu2017learning}                      & \cmark    & 1             & \xmark     & \cmark       & \cmark        & \xmark        & \cmark        & \xmark        & \xmark        & SL                & distance func.           \\
    \citet{yi2018deep}                          & \xmark    & 3             & \xmark     & \xmark       & \xmark        & \xmark        & \xmark        & \xmark        & \cmark        & SL                & displacement             \\    
    \citet{wang2019shape2motion}                & \xmark    & 1             & \xmark     & \cmark       & \cmark        & \xmark        & \xmark        & \xmark        & \xmark        & SL                & joint params.            \\
    RPM-Net~\cite{yan2019rpm}                   & \xmark    & arbitrary     & \xmark     & \cmark       & \cmark        & \xmark        & \xmark        & \xmark        & \cmark        & SL                & displacement             \\
    \citet{abbatematteo2019learning}            & \cmark    & 1             & \cmark     & \xmark       & \cmark        & \cmark        & \xmark        & \xmark        & \xmark        & SL                & joint params.            \\
    ANCSH~\cite{li2020category}                 & \cmark    & 1             & \cmark     & \cmark       & \cmark        & \cmark        & \xmark        & \xmark        & \xmark        & SL                & joint params.            \\
    CAPTRA~\cite{weng2021captra}                & \cmark    & 1             & \cmark     & \cmark       & \cmark        & \cmark        & \xmark        & \xmark        & \xmark        & SL                & joint params.            \\
    \citet{shi2021self}                         & \xmark    & 1             & \xmark     & \cmark       & \cmark        & \xmark        & \cmark        & \xmark        & \xmark        & SSL               & point corr.              \\
    ScrewNet~\cite{jain2021screwnet}            & \cmark    & 1             & \xmark     & \cmark       & \cmark        & \cmark        & \xmark        & \xmark        & \xmark        & SL                & joint params.            \\
    MultiBodySync~\cite{huang2021multibodysync} & \xmark    & 1             & \xmark     & \cmark       & \cmark        & \xmark        & \xmark        & \xmark        & \xmark        & SSL               & point corr.              \\
    \citet{abdul2022learning}                   & \xmark    & 1             & \xmark     & \cmark       & \xmark        & \xmark        & \xmark        & \cmark        & \xmark        & SL                & joint params.            \\
    \citet{qian2022understanding}               & \xmark    & 1             & \xmark     & \cmark       & \cmark        & \xmark        & \xmark        & \xmark        & \xmark        & SL                & joint params.            \\
    OPD~\cite{jiang2022opd}                     & \cmark    & 1             & \xmark     & \cmark       & \cmark        & \xmark        & \xmark        & \xmark        & \xmark        & SL                & joint params.            \\
    OPDMulti~\cite{sun2023opdmulti}             & \xmark    & 1             & \xmark     & \cmark       & \cmark        & \xmark        & \xmark        & \xmark        & \xmark        & SL                & joint params.            \\
    \citet{liu2022toward}                       & \xmark    & 1             & \xmark     & \cmark       & \cmark        & \cmark        & \xmark        & \cmark        & \xmark        & SL                & joint params.            \\
    \citet{liu2023self}                         & \cmark    & 1             & \cmark     & \cmark       & \cmark        & \cmark        & \xmark        & \xmark        & \xmark        & SSL               & point corr.              \\
    \citet{liu2023building}                     & \xmark    & 1             & \xmark     & \cmark       & \cmark        & \xmark        & \xmark        & \cmark        & \xmark        & SSL               & point corr.              \\
    GAPartNet~\cite{geng2023gapartnet}          & \xmark    & 1             & \xmark     & \cmark       & \cmark        & \cmark        & \xmark        & \xmark        & \xmark        & SL                & NPCS maps                \\
    \citet{liu2023semi}                         & \xmark    & 1             & \xmark     & \cmark       & \cmark        & \xmark        & \xmark        & \cmark        & \xmark        & SL                & joint params.            \\
    \jy{AutoURDF~\cite{lin2024autourdf}}        & \xmark    & 1             & \xmark     & \cmark       & \cmark        & \xmark        & \xmark        & \cmark        & \xmark        & SSL               & point corr.            \\
    GAMMA~\cite{yu2024gamma}                    & \xmark    & 1             & \xmark     & \cmark       & \cmark        & \xmark        & \xmark        & \xmark        & \xmark        & SL                & joint params.            \\

    \midrule \multicolumn{12}{l}{Articulated Object Creation} \\ \midrule 
    A-SDF~\cite{mu2021asdf}                     & \cmark    & 1             & \cmark     & \xmark       & \xmark        & \xmark        & \xmark        & \xmark        & \cmark        & SL                & joint params.            \\
    \citet{wei2022self}                         & \cmark    & 1             & \cmark     & \xmark       & \xmark        & \cmark        & \xmark        & \xmark        & \cmark        & SL                & joint params.            \\
    Ditto~\cite{jiang2022ditto}                 & \cmark    & 1             & \xmark     & \cmark       & \cmark        & \cmark        & \xmark        & \xmark        & \xmark        & SL                & joint params.            \\
    CLA-NeRF~\cite{tseng2022cla}                & \cmark    & 1             & \cmark     & \xmark       & \cmark        & \xmark        & \xmark        & \xmark        & \xmark        & handcrafted       & -                        \\
    CARTO~\cite{heppert2023carto}               & \cmark    & 1             & \xmark     & \cmark       & \xmark        & \cmark        & \xmark        & \xmark        & \xmark        & SL                & joint params.            \\
    \citet{liu2023few}                          & \cmark    & 1             & \cmark     & \xmark       & \xmark        & \cmark        & \xmark        & \cmark        & \xmark        & TL, handcrafted   & -               \\
    PARIS~\cite{liu2023paris}                   & \cmark    & 1             & \xmark     & \cmark       & \cmark        & \cmark        & \xmark        & \xmark        & \xmark        & SSL               & MV RGBs                  \\
    SfA~\cite{nie2023sfa}                       & \xmark    & 1             & \xmark     & \cmark       & \cmark        & \xmark        & \xmark        & \xmark        & \xmark        & SL                & labeled PC               \\
    NAP~\cite{lei2023nap}                       & \xmark    & 1             & \xmark     & \cmark       & \cmark        & \xmark        & \cmark        & \cmark        & \xmark        & generative        & joint params.                        \\
    CAGE~\cite{liu2024cage}                     & \xmark    & 1             & \cmark     & \cmark       & \cmark        & \xmark        & \cmark        & \cmark        & \xmark        & generative        & joint params.                        \\
    \citet{weng2024neural}                      & \cmark    & 1             & \xmark     & \cmark       & \cmark        & \cmark        & \xmark        & \xmark        & \xmark        & SSL               & MV RGB-Ds                \\
    REACTO~\cite{song2024reacto}                & \xmark    & arbitrary     & \xmark     & \xmark       & \xmark        & \xmark        & \xmark        & \xmark        & \cmark        & SL                & optical flow             \\
    Real2Code~\cite{mandi2024real2code}         & \xmark    & 1             & \xmark     & \cmark       & \cmark        & \xmark        & \cmark        & \xmark        & \xmark        & \jy{TL}           & joint params.            \\
    RSRD~\cite{kerr2024robot}                   & \xmark    & 1             & \xmark     & \xmark       & \xmark        & \cmark        & \xmark        & \xmark        & \xmark        & SSL               & MV RGBs, video            \\
    \jy{SINGAPO~\cite{liu2024singapo}}          & \xmark    & 1             & \xmark     & \cmark       & \cmark        & \xmark        & \cmark        & \cmark        & \xmark        & generative        & joint params.            \\
    \jy{Articulate-Anything~\cite{le2024articulate}} & \xmark    & 1        & \xmark     & \cmark       & \cmark        & \xmark        & \cmark        & \cmark        & \xmark        & \jy{TL}           & -            \\
    \jy{MeshArt~\cite{gao2024meshart}}               & \xmark    & 1        & \cmark     & \cmark       & \cmark        & \xmark        & \cmark        & \cmark        & \xmark        & generative        & joint params.            \\
    \jy{ArtFormer~\cite{su2024artformer}}           & \xmark    & 1         & \xmark     & \xmark       & \cmark        & \xmark        & \cmark        & \cmark        & \xmark        & generative        & joint params.            \\
    \jy{Articulate AnyMesh~\cite{qiu2025anymesh}}   & \xmark    & 1         & \xmark     & \cmark       & \cmark        & \xmark        & \cmark        & \cmark        & \xmark        & \jy{TL}, handcrafted & -                   \\
    ArtGS~\cite{liu2025building}                     & \xmark    & 1         & \xmark     & \cmark       & \cmark        & \cmark        & \xmark        & \xmark        & \xmark        & SSL               & MV RGB-Ds            \\
    \bottomrule
    \end{tabular}
    }
    \caption{
    Summary of the related works from the perspective of articulation modeling in terms of the assumptions about articulation structure each work makes, the representation of articulated motion, and the methodology used in the motion estimation process.
    }
    \label{tab:arti}
\end{table*}

%% file: figures/5_rpmnet.tex
\begin{figure}[t]
    \begin{center}
    \includegraphics[width=\linewidth]{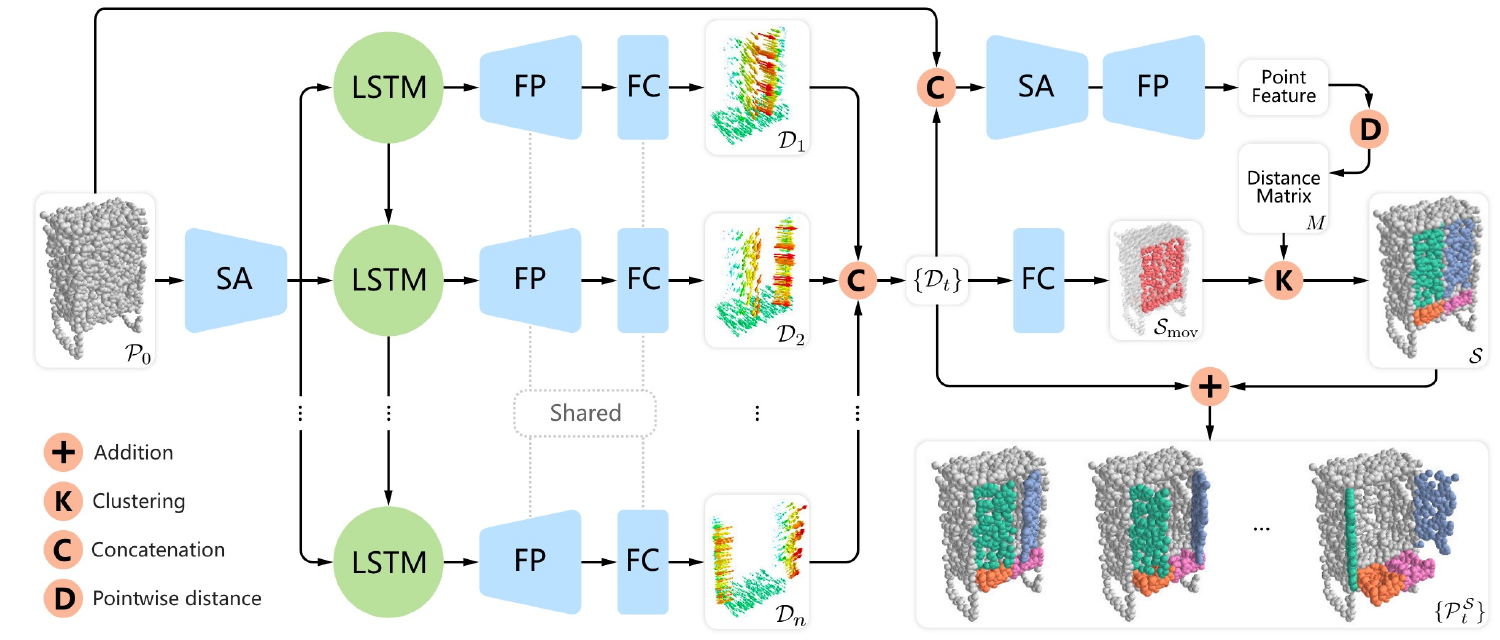}
    \end{center}
    \caption{An illustration of using a recurrent neural network to predict deformation field for each articulated part from RPMNet~\cite{yan2019rpm}.
    Figure reproduced from original paper~\cite{yan2019rpm}.}
    \label{fig:rpmnet}
\end{figure}

%% file: figures/5_acnsh.tex
\begin{figure}[t]
    \begin{center}
    \includegraphics[width=0.7\linewidth]{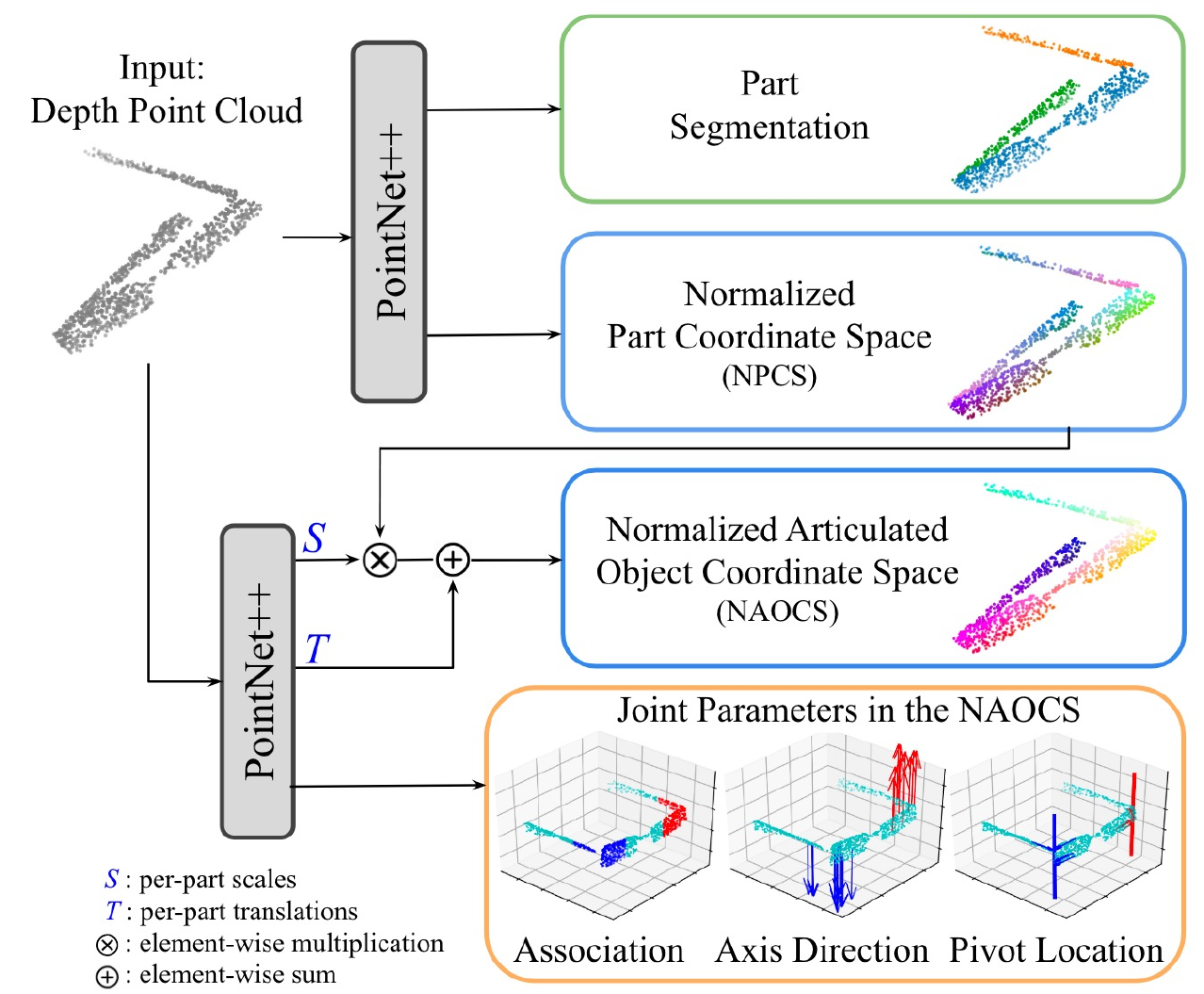}
    \end{center}
    \caption{Overview of the ANCSH method proposed by \citet{li2020category} that maps the object at an arbitrary articulated state to a hierarchical normalized coordinate space.
    Figure reproduced from original paper\cite{li2020category}.}
    \label{fig:acnsh}
\end{figure}

%% file: figures/5_paris.tex
\begin{figure}[t]
    \begin{center}
    \includegraphics[width=\linewidth]{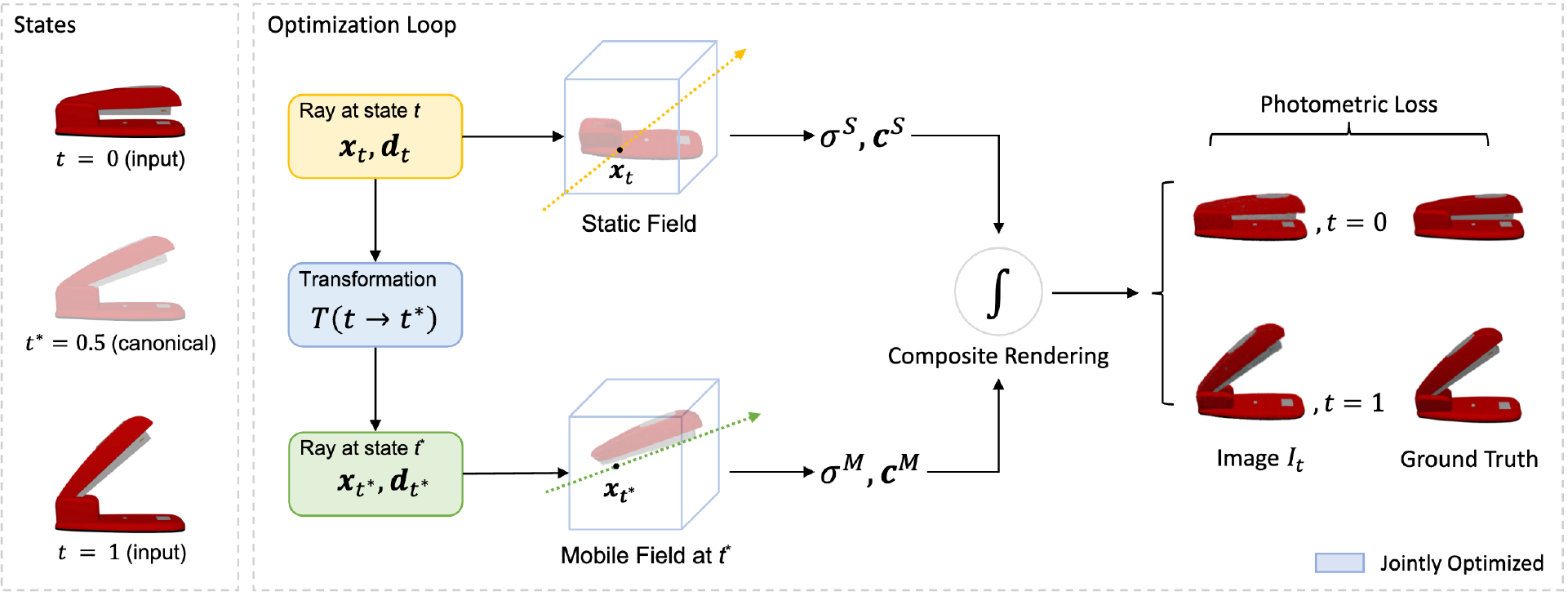}
    \end{center}
    \caption{An illustration of a self-supervised learning method proposed in PARIS~\cite{liu2023paris} that uses photometric consistency to jointly optimize part geometry and joint parameters .
    Figure reproduced from original paper~\cite{liu2023paris}.}
    \label{fig:paris}
\end{figure}

%% file: figures/5_articulate_anything.tex
\begin{figure}[t]
    \begin{center}
    \includegraphics[width=\linewidth]{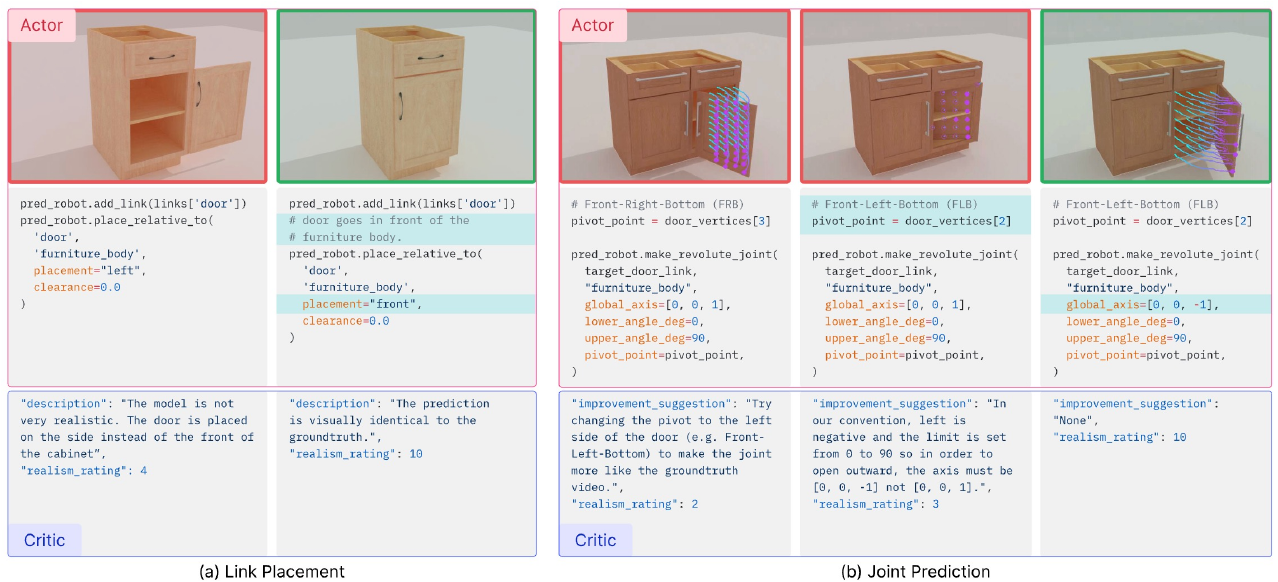}
    \end{center}
    \caption{The actor-critic program system for link placement and joint prediction proposed in Articulate-Anything~\cite{le2024articulate}. Figure reproduced from the original paper~\cite{le2024articulate}.}
    \label{fig:articulate_anything}
\end{figure}

%% file: secs/6_conclusion.tex
\input{figures/6_human_video.tex}

\section{Discussion and Conclusion}
\label{sec:conclusion}
 This survey paper provided an overview of the current state-of-the-art in 3D modeling of human-made articulated objects, focusing on the tasks of articulated part perception and articulated object creation (reconstruction and generation).
We first defined the scope of the discussion and provided a compilation of articulated object datasets used in the research community.
We then analyzed techniques and methodologies developed for 3D modeling of articulated objects, focusing on two key aspects: 1) \textit{geometry modeling} --- representation and approaches used to understand the shape of articulated parts and 2) \textit{articulation modeling} --- articulation models and methodologies used to estimate part mobility and kinematic structures.
We highlighted gaps in the literature and opportunities for further work emerging from our survey, which are likely to be fruitful as the field of 3D modeling for articulated objects continues to evolve.
We conclude by identifying key challenges and potential research directions for future exploration.

\mypara{Generalization across objects with varying structures}.
Handling diverse articulated objects with large variations in geometric and kinematic structures is one of the major challenges in the field.
Existing methods are often designed for specific object categories or assume a certain level of prior knowledge about the object is available to the system  (e.g., the number of parts, which joint types are considered).
Given limited training data and the high cost of data collection, it is important to develop methods that can generalize to novel objects in a data-efficient manner.
\jy{A promising direction is to explore the use of few-shot learning, meta-learning, and domain adaptation techniques to enhance the versatility of articulated object understanding.
Recent advances in large language models, vision-language models, and multimodal large language models can also be powerful prior to integrate into articulated object perception and manipulation pipelines, driving improvements in the generalization and robustness of the models.
}

\mypara{Improved robustness in the perception models}.
While significant progress has been made in articulated part perception, more accurate and robust methods are needed to handle occlusions, clutter, and noise in input data.
\jy{Most existing approaches focus on object-level perception with clear, object-centric observations, whereas perceiving articulated objects in complex scenes with multiple objects and partial observations remains an open challenge~\cite{delitzas2024scenefun3d}.
Addressing this gap is essential for making these methods practical and deployable in real-world applications, including functionality understanding, affordance learning, and robotic manipulation, where models must reliably interpret and interact with articulated objects in unstructured environments.}

\mypara{Detailed interior and part geometry modeling}.
Modeling interior and part-level geometry in detail for articulated objects  and further composing such detailed interactable objects at the scene level is an important aspect of geometric modeling that has not been fully explored.
For the object reconstruction task, the quality of the models is often limited by the quality of the input observation which is often partial due to view occlusion and sensor noise.
Developing methods that can handle these challenges and generate intricate and accurate 3D models of the articulated object  (e.g., fully functional cars, fridges with shelves and containers allowing small objects to be placed inside) is another promising direction for future research.

\mypara{Generative AI for articulated objects}.
\jy{The recent advances in articulated object generation have shown great potential in creating realistic and diverse 3D models of articulated objects.
Due to the highly structured nature and complexity in both geometry and articulation at the part level, generative models for articulated objects face unique challenges compared to general object generation.
The existing methods either prioritize structure modeling or geometry modeling, and the integration of both aspects while maintaining kinematic plausibility remains an open challenge.
Key challenges include designing efficient representations with compatible generative paradigms to capture the multi-faceted nature of articulated objects, and developing controllable models that can be prompted with different user inputs.
Additionally, data scarcity limits progress, as high-quality datasets with detailed part geometry and articulation annotations are rare, making it difficult to train scalable and generalizable models.
Promising future directions include expanding datasets, improving the representation learning, enhancing the multi-modal generation capabilities, and advancing controllable generative models for interactive design.
Another interesting venue is exploring domain-specific language models for articulated objects in combination with large language models.
The compositionality and the hierarchical structure of the articulated objects make them well suited for code- or program-based descriptions, and the integration of language priors can potentially improve the generation diversity and the model interpretability.
Ultimately, integrating generative AI with simulation environments will be essential for advancing interactive applications and embodied AI research, enabling more realistic modeling and interaction with everyday articulated objects.
}

\mypara{Incorporating physical constraints into the modeling process}.
Articulated objects are subject to physical constraints (e.g., collisions, materials) and physical properties (e.g., mass, inertia), which are often overlooked in existing modeling approaches.
\jy{Most current generative and perception models focus on capturing articulation geometry and motion patterns but do not explicitly enforce real-world physics, leading to unrealistic or non-functional outputs.}
Integrating the physical constraints of the objects can help to improve the accuracy of the articulation modeling \jy{and ensure that generated objects not only look realistic but also move and interact in a physically plausible manner}.
\jy{Considering physical constraints is particularly crucial for simulation-driven applications such as robotic manipulation, digital twins, and physics-based animation, where real-world interaction dynamics must be accurately replicated.}
This consideration is also useful in articulated object creation for applications such as virtual reality and gaming, \jy{where objects must react naturally to user interactions, maintaining realistic movements, and collision handling.}
\jy{Future research should explore integrating physics engines directly into the modeling pipeline to enforce kinematic and dynamic feasibility during object creation.
Additionally, learning-based approaches could benefit from differentiable physics simulation which allows models to be trained with explicit physics constraints, improving both realism and generalization. 
By embedding physical reasoning into the content creation process, we can bridge the gap between visual realism and functional accuracy, paving the way for more deployable and interactive articulated object models in robotics and virtual environments.}

\mypara{Leveraging human-in-the-loop data}.
Everyday human interactions with articulated objects produce a large amount of data in the form of long natural videos, egocentric videos, and human demonstrations (see examples in \Cref{fig:human_in_loop}).
These data are rich in articulated object information but also present challenges such as fast-moving objects, limited field-of-view camera viewpoints, streaming setups where the parts may disappear and reappear, and lots of occlusions and dynamics due to humans/hands in the view.
Collecting and using such data for articulated object modeling can enable generalization to diverse real objects and interactions.

\mypara{Evaluation and benchmarking}.
 Due to the multi-faceted nature of articulated objects, the evaluation of the quality of the created objects is currently not standardized.
The challenges of evaluation are partially due to the lack of ground truth data and the complexity of the task.
And even when the ground truth annotation is available, the metrics that have been proposed by prior work have flaws and are not fully and clearly specified.
There is a need for more systematic evaluation metrics that can evaluate different aspects of articulated object creation, including geometry, articulation, and part relationships, and also an object as a whole in the context of a scene where interaction can take place.

\mypara{Conclusion}.
This survey systematically reviewed the progress in 3D modeling for articulated objects, with a focus on articulated part perception and articulated object creation. 
By examining the development of articulated object modeling across the two axes of geometric and articulation modeling, we have identified significant advances and ongoing challenges in the field.
We also highlighted potential research directions that can drive future progress in the field.
We hope that this survey serves as a foundational reference for researchers and practitioners in computer vision and graphics, offering insights into the complexities of articulated object modeling and inspiring new research in this area.

%% file: figures/6_human_video.tex
\begin{figure*}[t]
    \begin{center}
    \includegraphics[width=\linewidth]{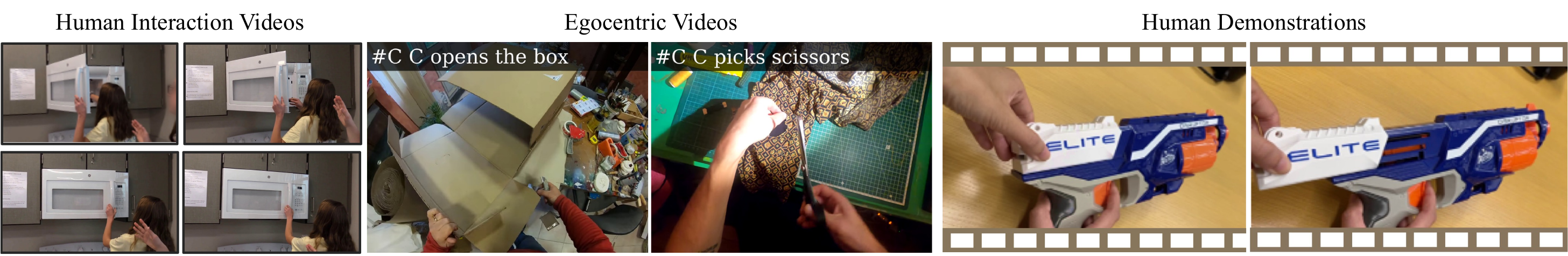}
    \end{center}
    \caption{ Example sources of human-in-the-loop video: human interactions, egocentric videos, and human demonstrations videos that involve articulated objects.
    The figures are reproduced from the original papers~\cite{qian2022understanding,grauman2022ego4d,kerr2024robot}.}
    \label{fig:human_in_loop}
\end{figure*}